\pdfoutput=1

\documentclass[11pt]{article}
\usepackage[table,dvipsnames]{xcolor}
\usepackage[preprint]{acl}

\usepackage{times}
\usepackage{latexsym}

\usepackage{multirow}
\usepackage{listings}
\usepackage{verbatim}
\usepackage{hhline}
\usepackage{tcolorbox}
\usepackage{enumitem}
\usepackage{afterpage}
\usepackage{placeins}
\usepackage{booktabs}
\definecolor{pygreen}{rgb}{0.0, 0.5, 0.0}
\definecolor{lightgray}{RGB}{211, 211, 211}
\definecolor{lightlightgray}{RGB}{211, 211, 211}
\definecolor{darkgray}{RGB}{100, 100, 100}

\usepackage{algorithm}
\usepackage{algpseudocode}
\usepackage{amsmath}
\usepackage{float}
\usepackage{rotating}  
\usepackage{array}     
\usepackage{longtable}
\newcolumntype{V}{!{\;\vrule width .6pt\;}}
\newcolumntype{R}[1]{>{\rotatebox[origin=c]{90}\arraybackslash}m{#1}}

\usepackage[T1]{fontenc}

\usepackage[utf8]{inputenc}

\usepackage{microtype}

\usepackage{inconsolata}

\usepackage{graphicx}

\title{ReaRAG: Knowledge-guided Reasoning Enhances Factuality of Large Reasoning Models with Iterative Retrieval Augmented Generation}

\author{ Zhicheng Lee$^1$, Shulin Cao$^1$, Jinxin Liu$^1$, Jiajie Zhang$^1$, Weichuan Liu$^2$,\\
\textbf{Xiaoyin Che$^2$, Lei Hou$^1$, Juanzi Li$^1$} \\
$^1$Tsinghua University, $^2$Siemens AG \\
\includegraphics[scale=0.035]{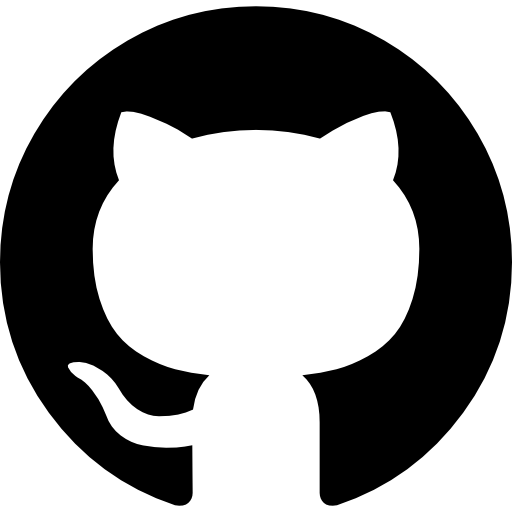} \url{https://github.com/THU-KEG/ReaRAG}
}

\begin{document}
\maketitle

\begin{abstract}
Large Reasoning Models (LRMs) exhibit remarkable reasoning abilities but rely primarily on parametric knowledge, limiting factual accuracy. While recent works adopt reinforcement learning (RL) training to integrate reasoning with retrieval, such methods are often complex and resource-heavy. Instead, our approach achieves comparable performance via strategic distillation, without the need for costly RL training. In this paper, we propose \textbf{ReaRAG}, a factuality-enhanced reasoning model that explores diverse queries without excessive iterations. Our solution includes a novel data construction framework with an upper bound on the reasoning chain length. Specifically, we first leverage an LRM to generate deliberate thinking, then select an action from a predefined action space (\texttt{Search} and \texttt{Finish}). For \texttt{Search} action, a query is executed against the RAG engine, where the result is returned as observation to guide reasoning steps later. This process iterates until a \texttt{Finish} action is chosen. Benefiting from ReaRAG's strong reasoning capabilities, our approach outperforms existing baselines overall. Further analysis highlights its strong reflective ability to recognize errors and refine its reasoning trajectory. Our study enhances LRMs’ factuality while effectively integrating robust reasoning for Retrieval-Augmented Generation (RAG).
\end{abstract}


\section{Introduction}
Large Reasoning Models (LRMs) such as OpenAI's o1 \citep{DBLP:journals/corr/abs-2412-16720}, Qwen's QwQ-32B\footnote{For simplicity, all mentions of QwQ-32B in this paper refer to QwQ-32B-Preview.} \citep{qwen-qwq}, GLM-Z1\footnote{\label{glm-z1-footnote}\url{https://open.bigmodel.cn/dev/api/Reasoning-models/glm-z1}} and DeepSeek-R1 \citep{DBLP:journals/corr/abs-2501-12948} demonstrate impressive reasoning capabilities on complex tasks \citep{DBLP:journals/corr/abs-2501-09686}. However, their reliance on parametric knowledge during reasoning limits performance on question answering (QA) tasks that require factual answers, where reasoning beyond the model's parametric knowledge is required.


To enhance LRMs' factuality, Retrieval-Augmented Generation (RAG) \citep{DBLP:conf/nips/LewisPPPKGKLYR020, DBLP:conf/naacl/ShiMYS0LZY24, DBLP:conf/icml/GuuLTPC20} offers a promising solution by integrating external knowledge but faces challenges in retrieving relevant documents, which requires formulating precise search queries \citep{DBLP:journals/corr/abs-2404-00610}. Prior research has explored iterative retrieval strategies \citep{DBLP:conf/emnlp/PressZMSSL23, DBLP:conf/emnlp/ShaoGSHDC23}, which construct reasoning chains of sub-queries and sub-answers to solve multi-hop QA. However, these methods suffer from error propagation, where mistakes in earlier steps mislead subsequent retrieval and reasoning, ultimately degrading the overall answer quality \citep{DBLP:conf/emnlp/CaoZSL0THL23}. 


\begin{figure}
 \centering
  \includegraphics[width=0.9\linewidth]{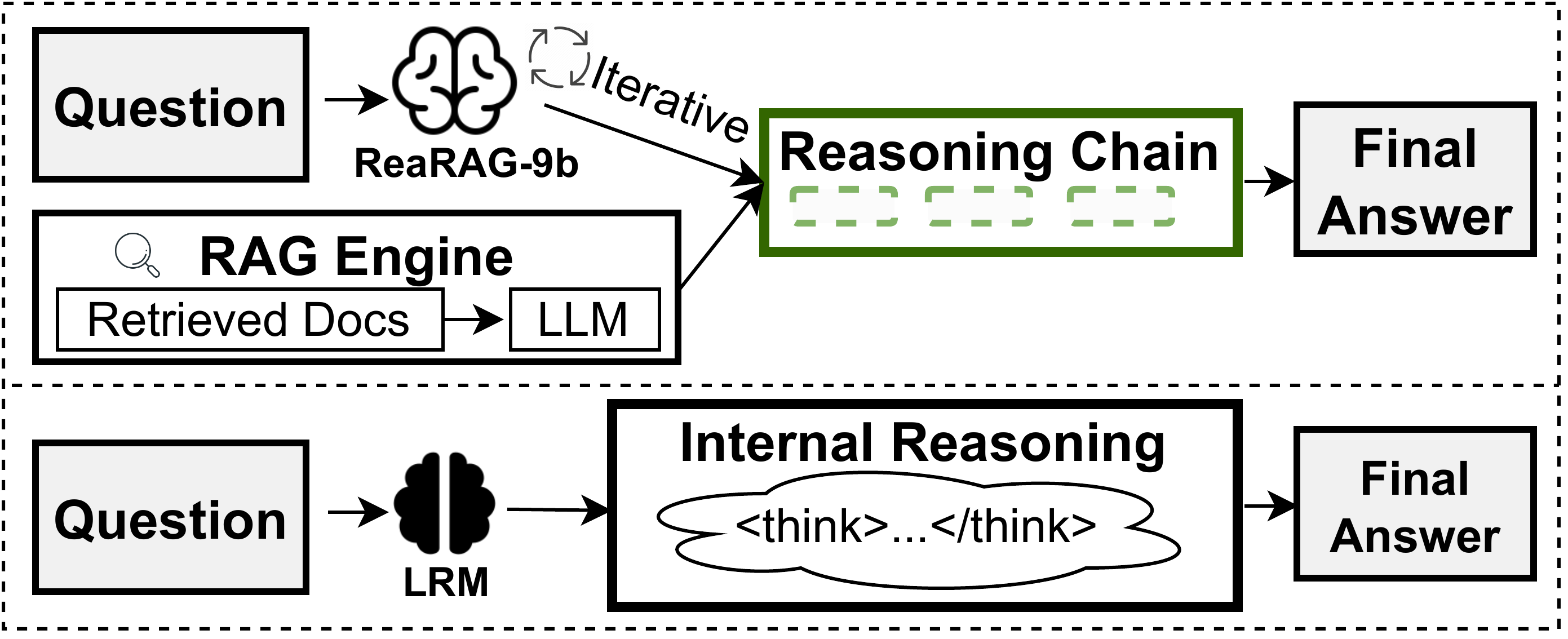}
  \caption {Unlike LRMs, ReaRAG iteratively constructs knowledge-guided reasoning chains for factual answers.}
  \label{fig:intro}
\end{figure}

To address this limitation, Search-o1 \citep{DBLP:journals/corr/abs-2501-05366} prompts a LRM to iteratively retrieve documents and generate sub-answers via a Reason-in-Documents module, relying heavily on the model’s inherent reasoning and instruction-following capabilities. This leads to three key issues: (1) unreliable retrieval token generation, (2) information extraction failures and hallucinations during document reasoning, and (3) overthinking in multi-hop QA \citep{DBLP:journals/corr/abs-2412-21187}. Effectively integrating reasoning with retrieval remains challenging.

In response to this challenge, recent works such as R1-Searcher \citep{DBLP:journals/corr/abs-2503-05592} and ReSearch \citep{DBLP:journals/corr/abs-2503-19470} adopt reinforcement learning (RL) to incentivize Large Language Models (LLMs) to perform reasoning before retrieval. However, such policy optimization approaches involve complex training paradigms and require extensive computational resources.

In this paper, we propose \textbf{ReaRAG}, a factuality-enhanced \textbf{rea}soning model for \textbf{RAG} that achieves comparable performance without relying on complex RL-based training. Instead of resource-heavy policy optimization, we strategically distill the reasoning capabilities of LRMs to iteratively construct knowledge-guided reasoning chains, which are then curated into a dedicated dataset with restricted chain lengths to fine-tune ReaRAG under the Thought-Action-Observation paradigm. During inference, ReaRAG iteratively performs the \texttt{search} action and dynamically decides when to trigger the \texttt{finish} action, avoiding excessive retrieval. Guided by external knowledge, ReaRAG reflects on its reasoning trajectory, detects errors, and realigns its reasoning toward the correct path, leading to improved performance with greater training efficiency and practical implementation.

To validate our approach, we conduct experiments on six benchmarks covering both multi-hop and single-hop QA, where ReaRAG achieves the highest overall performance. In summary, our contributions are as follows:


\begin{itemize}[wide, labelwidth=!, topsep=2pt]
\item \textbf{Enhancing LRMs' factuality through knowledge-guided reasoning chain}. We propose ReaRAG-9B, a model fine-tuned on a dedicated dataset to perform knowledge-guided reasoning, enabling reliable access to external knowledge. Analyses show that ReaRAG reflects on prior steps, strategically uses external knowledge to identify errors and refines its reasoning, showcasing robust multi-step reasoning capabilities.

\item  \textbf{Effectively combining strong reasoning with RAG}. 
Compared to Search-o1, our model avoids redundant retrieval in multi-hop QA. Compared to RL-based methods, we offer a more efficient and robust solution by strategically distilling reasoning capabilities from a LRM, achieving comparable performance even without complex RL training.


\item \textbf{Enhanced benchmark performance}. Compared to RL-based methods, ReaRAG performs on par with or surpasses them on MuSiQue \citep{DBLP:journals/tacl/TrivediBKS22}, HotpotQA \citep{DBLP:conf/emnlp/Yang0ZBCSM18}, IIRC \citep{DBLP:conf/emnlp/FergusonGHKD20} and Natural Questions (NQ) \citep{DBLP:journals/tacl/KwiatkowskiPRCP19}, while outperforming them on harder tasks such as FRAMES \cite{DBLP:journals/corr/abs-2409-12941} and FanOutQA \citep{DBLP:conf/acl/ZhuHDC24}.


\end{itemize}

\section{Related Work}
\begin{figure*}[t]
 \centering
  \includegraphics[width=0.9\linewidth]{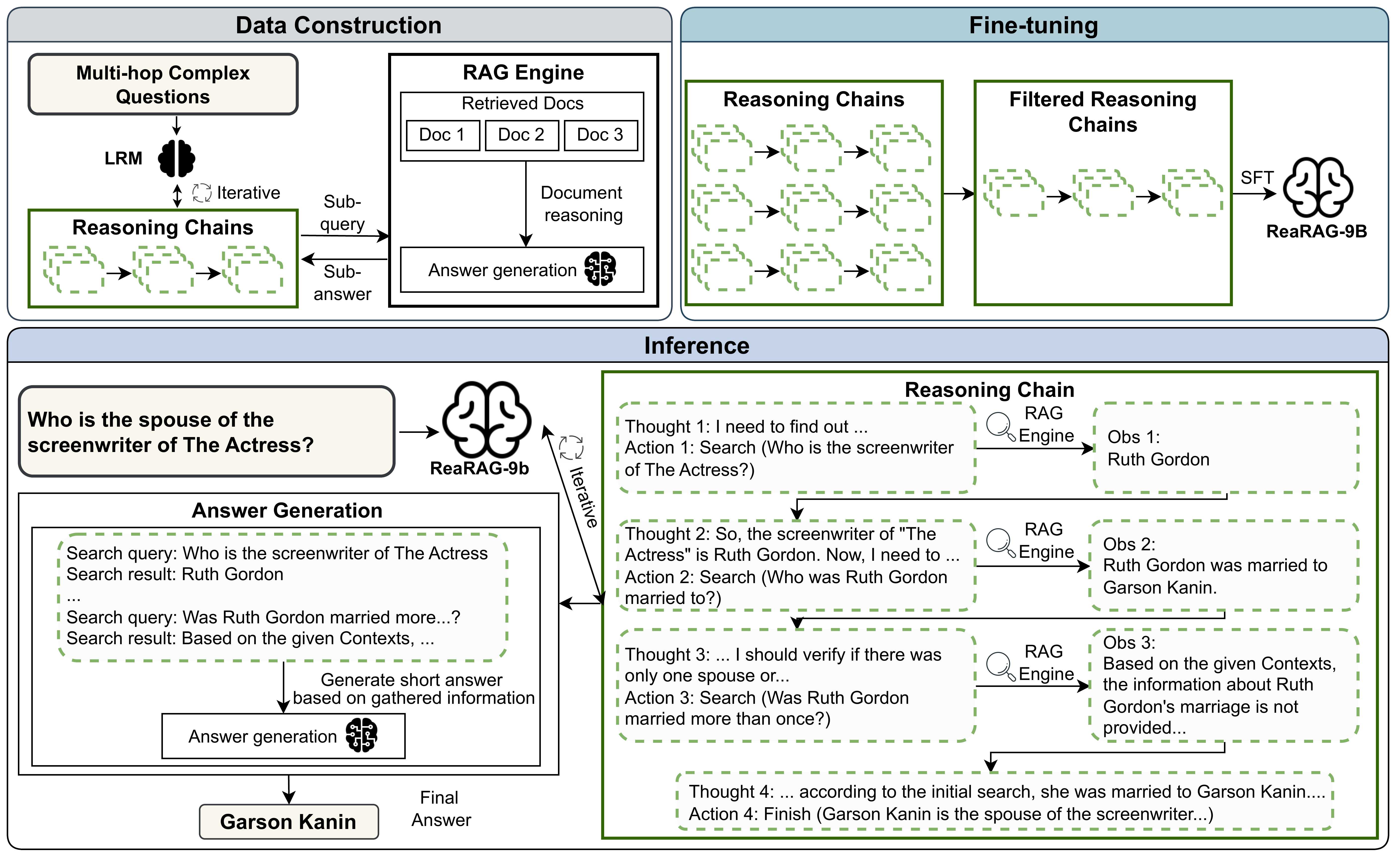}
  \caption {Overview of our approach to develop a factuality-enhanced reasoning model \textbf{ReaRAG}. To equip ReaRAG with knowledge-guided reasoning ability, we propose an automated data construction approach (Algorithm~\ref{alg:data_construction}). Next, we fine-tune ReaRAG on the constructed dataset to conduct reasoning iteratively, following the Thought-Action-Observation Paradigm to solve complex queries. Pseudocode for the inference stage is provided in Algorithm~\ref{alg:inference}.}
  \label{fig:overview}
\end{figure*}
\paragraph{Reasoning-enhanced LLMs.} Numerous works have investigated how to elicit the reasoning abilities of LLMs. Early approaches, such as Chain-of-Thought (COT) \citep{DBLP:conf/nips/Wei0SBIXCLZ22}, ReAct \citep{DBLP:conf/iclr/YaoZYDSN023} and Tree of Thought (ToT) \citep{DBLP:conf/nips/YaoYZS00N23} prompts LLMs to generate human-like step by step reasoning chain, though they still struggle on complex reasoning tasks. Recent advances of LRMs have scaled up CoT through RL \citep{kumar2024training}, enabling models to generate long CoT before providing a final answer. Notably, LRMs include OpenAI's o1 \citep{DBLP:journals/corr/abs-2412-16720}, Qwen's QwQ-32B \citep{qwen-qwq}, GLM-Z1 and DeepSeek-R1\citep{DBLP:journals/corr/abs-2501-12948} demonstrate impressive performance in complex reasoning tasks. Despite these advancements, most LRMs lack the ability to interact with external knowledge, limiting their capacity to generate factual responses.

\paragraph{Retrieval-Augmented Generation.} RAG has emerged as a promising paradigm for improving LLMs' factuality. Early methods rely on a single retrieval step \citep{DBLP:conf/nips/LewisPPPKGKLYR020, DBLP:conf/icml/BorgeaudMHCRM0L22, DBLP:journals/jmlr/IzacardLLHPSDJRG23}, which often falls short for multi-hop QA tasks due to limited retrieval quality. To mitigate noisy retrieval \citep{DBLP:conf/icml/ShiCMSDCSZ23}, Self-RAG \citep{DBLP:conf/iclr/AsaiWWSH24} and CRAG \citep{DBLP:journals/corr/abs-2401-15884} introduce reflection mechanisms on the retrieved documents, yet single-step retrieval remains insufficient. To address these limitations, iterative methods such as Iter-RetGen \citep{DBLP:conf/emnlp/ShaoGSHDC23} and Self-Ask \citep{DBLP:conf/emnlp/PressZMSSL23} progressively retrieve relevant documents to gather sufficient information for final answer. Building on this, SearChain \citep{DBLP:conf/www/XuPSCC24} first generates a complete reasoning chain, then verifies the answer of each node in the chain based on external knowledge. However, these approaches lack a strong reflection mechanism to recover from earlier reasoning errors.

\paragraph{Reasoning-enhanced RAG.} Recent studies in scaling token generation at test-time \citep{learning-to-reason, muennighoff2025s1, DBLP:journals/corr/abs-2408-03314} to enhance LLMs' reasoning capabilities have spurred interest in reasoning-enhanced RAG. RAG-Star \citep{DBLP:journals/corr/abs-2412-12881} leverages Monte Carlo Tree Search to iteratively decompose multi-hop questions, guided by a reward model during the tree
expansion. Search-o1 \citep {DBLP:journals/corr/abs-2501-05366} introduces a document reasoning module and prompts QwQ-32B to generate special tokens that trigger knowledge retrieval. However, it heavily depends on the base model's instruction-following and inherited reasoning capabilities, leading to three key challenges: unreliable generation of specialized tokens, failure in extracting information from retrieved documents, and overthinking for multi-hop QA. To integrate reasoning with retrieval, R1-Searcher \citep{DBLP:journals/corr/abs-2503-05592} and ReSearch \citep{DBLP:journals/corr/abs-2503-19470} adopt a RL-based method that incentivizes LLMs to reason before retrieving. However, despite the complexity of RL-based training paradigms, our approach achieves comparable performance through strategic distillation, offering greater training efficiency and practical ease of implementation.


\section{Methodology}
\label{sec:method}

\begin{algorithm}[t]
\caption{Data Construction}
\label{alg:data_construction}
\textbf{Input:} Seed Dataset $\mathcal{D_{\text{seed}}}$, Large Reasoning Model $\mathcal{M}_{LRM}$, Instruction prompt $\mathcal{P}_{d}$, Max iterations $T_{\text{max}}$, RAG engine $\mathcal{R}$ \\
\textbf{Output:} Dataset $\mathcal{D}_{\text{reason}}$ with reasoning chains
\begin{algorithmic}[1]
\State Initialize $\mathcal{D}_{\text{reason}} \gets \emptyset$
\For{each $(x_i, doc_i) \sim \mathcal{D_{\text{seed}}}$}
    \Statex \Comment{Sample question and gold documents}
    \State $t \gets 0$ \Comment{Iteration counter}
    \State $\mathcal{C}_i \gets []$ \Comment{Reasoning chain}
    \While{$t < T_{\text{max}}$}
        \State $y'_t \gets \mathcal{M}_{LRM}([\mathcal{P}_{d} \oplus x_i \oplus \mathcal{C}_i])$
        \Statex \Comment{Generate response}
        \State $(\tau_t, \alpha_t) \gets \texttt{parse}(y'_t)$ 
        \Statex \Comment{Extract thought $\tau_t$ and action $\alpha_t$}
        \State Get action type $\alpha_{t_{\text{type}}} \gets \alpha_t[\text{`type'}]$
        \If{$\texttt{finish} \in \alpha_{t_{\text{type}}}$}
            \State Append $(\tau_t, \alpha_t)$ to $\mathcal{C}_i$
            \State \textbf{break} 
        \ElsIf{$\texttt{search} \in\alpha_{t_{\text{type}}}$}
            \State $q_s \gets \alpha_t[\text{`query'}]$
            \State $o_t \gets \mathcal{R}(q_s, doc_i)$ 
            \Statex \Comment{Get $o_t$ from RAG engine}
            \State Append $(\tau_t, \alpha_t, o_t)$ to $\mathcal{C}_i$
        \EndIf
        \State $t \gets t + 1$
    \EndWhile
    \State Append $\mathcal{C}_i$ to $\mathcal{D}_{\text{reason}}$
\EndFor
\State \textbf{return} $\mathcal{D}_{\text{reason}}$
\end{algorithmic}
\end{algorithm}

In this section, we first formalize the task, then present our novel approach for developing ReaRAG as illustrated in Figure~\ref{fig:overview}.


\subsection{Task formulation}
We focus on the multi-hop QA task, where iterative reasoning improves answer accuracy. Given a question $x$, our goal is to construct a knowledge-guided reasoning chain $\mathcal{C}$ to enhance the factual correctness of the generated answer $\hat{y}$. Specifically, the reasoning chain is formulated as a sequence of $N$ steps, where each step consists of a reasoning thought $\tau_t$, an action $\alpha_t$, and an observation $o_t$:

\begin{equation}
  \label{eq:reasoning_chain}
  \mathcal{C} = \{ (\tau_t, \alpha_t, o_t) \}_{t=1}^{N}, \quad 1\leq t\leq T_{max}
\end{equation}

The number of reasoning steps $N$ is dynamically determined by the model but is constrained by an upper limit $T_{max}$ to prevent indefinite iterations, i.e., $N \leq T_{max}$. To guide the reasoning process with external knowledge, we define the action space as $\mathcal{A} = \{{\texttt{search()}, \texttt{finish()}}\}$ where \texttt{search} takes a search query as input, while the input to \texttt{finish} represents the derived answer. At each step $t$, the model refines a search query based on the reasoning thought $\tau_t$ and executes the $\texttt{search}$ action to retrieve relevant information from the RAG engine $\mathcal{R}$. The process continues until the model selects the $\texttt{finish}$ action, at which point the final answer $\hat{y}$ is derived from all prior reasoning steps. This ensures that the answer is grounded in retrieved knowledge through an iterative and structured reasoning process, thereby enhancing the factual reliability of LRMs.




\subsection{Knowledge-guided reasoning chain generation}
While existing LRMs demonstrate strong reasoning capabilities, they often fail to ground their reasoning process in factual knowledge. To make external knowledge accessible, we design a structured reasoning step where each step consists of a reasoning thought $\tau_t$, an action $\alpha_t$, and an observation $o_t$. 

\begin{itemize}[wide, labelwidth=!, topsep=2pt]
    \item Reasoning thought $\tau_t$: Represents the model's thought process where it reflects on prior actions and observations before deciding an action and its input parameter.
    \item Action $\alpha_t$: A JSON dictionary containing an action sampled from the action space $\mathcal{A}$ along with the corresponding input parameter. 
    \item Observation $o_t$: Feedback received after executing action $\alpha_t$, guiding subsequent reasoning.
\end{itemize}

To equip ReaRAG with the ability to construct reasoning chain guided by external knowledge, we propose an automated data construction approach, as detailed in Algorithm~\ref{alg:data_construction}.

\paragraph{Data construction.} Given a multi-hop question $x_i$ sampled from the seed dataset, we prompt the LRM $\mathcal{M}_{LRM}$ with instruction prompt $\mathcal{P}_{d}$ (see Appendix~\ref{sec:prompts}) to collect the reasoning thoughts and actions. Next, the search query is extracted and executed against RAG Engine $\mathcal{R}$ to obtain an observation $o_t$. The process iterates until either model decides an $\texttt{finish}$ action or the iteration count exceeds the maximum iterations $T_{max}$.

\paragraph{Data filtering.} Previous work studies have shown that the performance of the LLMs heavily depends on the quality of fine-tuning data \citep{DBLP:journals/corr/abs-2306-11644}. To ensure high-quality reasoning chains, we apply data filtering by comparing the final answer $\hat{y}_i$ derived from the reasoning chain $\mathcal{C}_i$ against the ground truth answer $y_i$ using the F1 metric. Reasoning chains with an F1 score of $0$ are discarded to maintain data integrity.

\subsection{Factuality-enhanced LRMs: ReaRAG}
\paragraph{Fine-tuning.} To incorporate knowledge-guided reasoning ability into the model, we perform supervised fine-tuning (SFT) on the constructed dataset discussed in the previous section. Each sample is a sequence of conversation chain $\mathcal{S} = \{\mathcal{P}, x_i, \{ (\tau_t, \alpha_t, o_t) \}_{t=1}^{N}\}$, where $\mathcal{P}$ denotes the instruction prompt (see Appendix~\ref{sec:prompts}). We fine-tune our factuality-enhanced LRM, ReaRAG ($\mathcal{M}_{ReaRAG}$) using loss function below:

\begin{equation}
    L = - \sum_{j} \mathbf{1} \times \log \mathcal{M}_{ReaRAG}(s_j \mid s_{<j})
\end{equation}

where $s_j$ represents the textual tokens of the input sequence $\mathcal{S}$, $\mathbf{1}(\cdot)$ is a loss mask indicator, which is set to True on thought $\tau_t$ and action $\alpha_t$ tokens, ensuring that the loss is only computed over the tokens contributed to the thought reasoning as well as action, rather than the entire sequence.

\paragraph{Inference.} After fine-tuning, ReaRAG is equipped with advanced reasoning capabilities to solve multi-hop QA. Given an instruction prompt $\mathcal{P}$ and a question $x$, the model first generates a reasoning thought $\tau_0$ and an initial action $\alpha_0$, typically a $\texttt{search}$ action. The search query $q_s$ is extracted and executed over the RAG engine $\mathcal{R}$, which returns an observation $o_0$. The triplet $(\tau_t, \alpha_t, o_t)$ is appended to the reasoning chain $\mathcal{C}$, while the pair $(q_s, o_t)$ is appended to the summary chain $Sum$. This process iterates until ReaRAG decides on a $\texttt{finish}$ action at step $N$. Upon this action, we concatenate all gathered information in $Sum$ and prompt an answer model $\mathcal{M}_{Ans}$ to generate final answer $\hat{y}$ using the prompt $\mathcal{P}_{ans}$ described in Appendix~\ref{sec:prompts}. The pseudocode for the inference stage is provided in Algorithm~\ref{alg:inference}. 

\section{Experiments}

\begin{algorithm}[t]
\caption{Inference}
\label{alg:inference}
\textbf{Input:} Input question $x$, documents $doc$, ReaRAG $\mathcal{M}_{ReaRAG}$, Answer LLM $\mathcal{M}_{Ans}$, Instruction prompt $\mathcal{P}$, Answer Prompt $\mathcal{P}_{ans}$, Max iterations $T_{\text{max}}$, RAG engine $\mathcal{R}$ \\
\textbf{Output:} Final answer, $\hat{y}$
\begin{algorithmic}[1]

\State $t \gets 0$, $\mathcal{C} \gets []$ $Sum \gets []$ \Comment{Initialization}


\While{$t < T_{\text{max}}$}
    \State $y'_t \gets \mathcal{M}_{ReaRAG}([\mathcal{P} \oplus x \oplus \mathcal{C}])$
    \Statex \Comment{Generate response for iteration $t$}
    \State $(\tau_t, \alpha_t) \gets \texttt{parse}(y'_t)$ 
    \Statex \Comment{Extract thought $\tau_t$ and action $\alpha_t$}
    \State Get action type $\alpha_{t_{\text{type}}} \gets \alpha_t[\text{`type'}]$
    \If{$\texttt{finish} \in \alpha_{t_{\text{type}}}$}
        \State $\hat{y} \gets \mathcal{M}_{Ans}([ \mathcal{P}_{ans} \oplus x \oplus Sum])$
        \State \textbf{return} final answer $\hat{y}$
    \ElsIf{$\texttt{search} \in\alpha_{t_{\text{type}}}$}
        \State $q_s \gets \alpha_t[\text{`query'}]$
        \State $o_t \gets \mathcal{R}(q_s, doc)$ 
        \Statex \Comment{Get $o_t$ from RAG engine}
        \State Append $(\tau_t, \alpha_t, o_t)$ to $\mathcal{C}$
        \State Append $(q_s, o_t)$ to $Sum$
    \EndIf
    \State $t \gets t + 1$
\EndWhile
\end{algorithmic}
\end{algorithm}



\begin{table*}[t]
    \centering
    \resizebox{1\linewidth}{!}{
    \begin{tabular}{l *{10}{c} V *{2}{c} V c}
    \toprule
    \multirow{3}{*}{\textbf{Model}} 
    & \multicolumn{10}{c V}{\textbf{Multi-hop}}
    & \multicolumn{2}{c V}{\textbf{Single-hop}}
    & \multirow{3}{*}{\textbf{Average}} \\
    \cmidrule(r){2-11}\cmidrule(l){12-13}
     & \multicolumn{2}{c}{\textbf{MuSiQue}} 
     & \multicolumn{2}{c}{\textbf{HotpotQA}} 
     & \multicolumn{2}{c}{\textbf{IIRC}} 
     & \multicolumn{2}{c}{\textbf{FRAMES}} 
     & \multicolumn{2}{c V}{\textbf{FanOutQA}}
     & \multicolumn{2}{c V}{\textbf{NQ}} \\
    \cmidrule(lr){2-3}\cmidrule(lr){4-5}\cmidrule(lr){6-7}\cmidrule(lr){8-9}\cmidrule(lr){10-11}\cmidrule(lr){12-13}\cmidrule(l){14-14}
     & $\text{ACC}_{L}$ & EM 
     & $\text{ACC}_{L}$ & EM 
     & $\text{ACC}_{L}$ & EM 
     & $\text{ACC}_{L}$ & EM 
     & $\text{ACC}_{L}$ & R-L 
     & $\text{ACC}_{L}$ & EM
     & $\text{ACC}_{L}$ \\
    \midrule
    \multicolumn{14}{l}{\textit{\textbf{In-context}}} \\
    GLM-4-9B (128k) & 23.5 & 15.0 & 58.0 & 47.0 & - & - & - & - & - & - & 45.5 & 26.0 & - \\
    GLM-4-32B (128k) & 33.5 & 17.0 & 65.5 & 50.0 & - & - & - & - & - & - & 52.5 & 24.0 & - \\
    \midrule
    \multicolumn{14}{l}{\textit{\textbf{Vanilla RAG}}} \\
    GLM-4-9B (128k) & 25.5 & 14.0 & 68.0 & 52.0 & 27.3 & 22.0 & 11.0 & 7.3 & 4.2 & 15.6 & 49.0 & 32.0 & 30.8 \\
    GLM-4-32B (128k) & 29.0 & 17.0 & 67.5 & 52.0 & 26.0 & 14.5 & 13.9 & 7.6 & 3.9 & 12.3 & \underline{53.0} & \textbf{39.0} & 32.2 \\
    QwQ-32B          & 36.0 & 20.0 & 67.0 & 47.0 & 32.8 & 25.0 & 23.4 & 13.9 & 9.7 & 14.2 & 48.0 & 26.0 & 36.1\\
    \midrule
    \multicolumn{14}{l}{\textit{\textbf{Advanced RAG}}} \\
    Self-RAG-7B   & 24.0 & 13.0 & 45.5 & 31.0 & 21.8 & 12.0 & 8.8 & 5.6 & 3.9 & 14.2 & 40.0 & 28.0 & 24.0 \\
    SearChain-7B     & 8.5 & 3.0 & 23.0 & 17.0 & 25.8 & 18.5 & 16.0 & 10.6 & 4.8 & \underline{17.7} & 28.0 & 13.0 & 17.7 \\
    Search-o1-32B    & 40.5 & 32.0 & 55.5 & 38.0 & 32.8 & 29.5 & 27.8 & 19.2 & \textbf{12.3} & 14.9 & 43.0 & 28.0 & 35.3 \\
    R1-Searcher-7B-Base    & 64.0 & \textbf{52.5} & \underline{79.5} & \textbf{68.0} & \textbf{42.5} & \textbf{35.5} & 26.9 & 18.7 & 5.8 & 15.6 & 48.5 & 35.0 & 44.5 \\
    ReSearch-Qwen-7B-Instruct    & \underline{65.0} & \underline{49.0} & 72.5 & \underline{59.0} & \underline{41.8} & \underline{33.5} & \underline{29.4} & \underline{21.0} & 6.8 & 15.8 & \underline{53.0} & \underline{37.0} & \underline{44.7} \\
    \midrule
    \multicolumn{14}{l}{\textit{\textbf{Ours}}} \\
    \textbf{ReaRAG-9B} & \textbf{67.0} & 46.0 & \textbf{81.5} & 58.0 & 41.0 & 27.5 & \textbf{34.2} & \textbf{21.2} & \underline{11.6} & \textbf{24.3} & \textbf{55.5} & 32.0 & \textbf{48.5} \\
    \bottomrule
    \end{tabular}
    }
    \caption{\label{tab:main-experiment}
        Main experimental results across six benchmarks. \textbf{Bold} and \underline{underline} indicate the best and second best results. We report the traditional EM scores as well as $\text{ACC}_{L}$, a metric based on the LLM-as-a-Judge framework using GPT-4o. For FanOutQA, we follow the original benchmark and report the Rouge-L (R-L) score. Our model, \textbf{ReaRAG-9B}, achieves the best overall performance, demonstrating the effectiveness of our strategic distillation approach, even when compared to recent RL-based baselines that integrate reasoning with retrieval.}
\end{table*}

\subsection{Experimental setup}


\paragraph{Dataset and metrics.} We evaluate our approach on multi-hop QA benchmarks requiring reasoning across multiple documents, including MuSiQue (MQ) \citep{DBLP:journals/tacl/TrivediBKS22}, HotpotQA (HP) \citep{DBLP:conf/emnlp/Yang0ZBCSM18} and IIRC \citep{DBLP:conf/emnlp/FergusonGHKD20}, as well as recently proposed benchmarks, including FRAMES (FR) \cite{DBLP:journals/corr/abs-2409-12941} and FanOutQA (FQ) \citep{DBLP:conf/acl/ZhuHDC24}. To evaluate single-hop reasoning capabilities, we additionally include NQ \citep{DBLP:journals/tacl/KwiatkowskiPRCP19}. Since these datasets require open-ended answers, traditional metric such as exact match (EM) may fail to capture semantically equivalent responses \citep{DBLP:conf/acl/YinBRCC0L24}. Hence, we include the LLM-as-a-Judge metric ($\text{ACC}_L$) \citep{DBLP:conf/nips/ZhengC00WZL0LXZ23} based on GPT-4o for more accurate evaluation using the prompt $\mathcal{P}_{judge}$ described in Appendix~\ref{sec:prompts}. For FanOutQA, we follow the metric used in the original benchmark and report the Rouge-L (R-L) score instead of EM. We randomly sample 100 validation examples each from MuSiQue, HotpotQA, and NQ, 200 from IIRC, and 396 from FRAMES. For FanOutQA, we evaluate on the full development set, which contains 310 examples.

\paragraph{Baselines.} We compare our approach against multiple baselines, categorized based on their access to external knowledge. These include \textit{in-context} retrieval, where the corpus is directly appended to the language model's context; \textit{vanilla RAG}, which performs a single retrieval based on the original input question; and state-of-the-art \textit{advanced RAG} methods proposed recently.

In the \textit{in-context} settings, we use GLM-4-9B and GLM-4-32B \citep{DBLP:journals/corr/abs-2406-12793}, both with a 128k context window. The \textit{vanilla RAG} baselines employ the same models, along with QwQ-32B \citep{qwen-qwq}. For \textit{advanced RAG} baselines, we include Self-RAG \citep{DBLP:conf/iclr/AsaiWWSH24}, which fine-tunes Llama-2-7B to retrieve documents on demand and filter noisy evidence; SearChain \citep{DBLP:conf/www/XuPSCC24}, which generates a Chain-of-Query structure enabling multi-turn retrieval with verification; and Search-o1 \citep{DBLP:journals/corr/abs-2501-05366}, which proposes a framework for LRMs to perform iterative knowledge retrieval. We further consider R1-Searcher \citep{DBLP:journals/corr/abs-2503-05592} and ReSearch \citep{DBLP:journals/corr/abs-2503-19470}, which incentivize LLMs to perform reasoning prior to retrieval through RL.

\subsection{Implementations details}

\paragraph{Data construction and fine-tuning.} The seed dataset described in Algorithm~\ref{alg:data_construction} is derived from the training sets of MuSiQue, HotpotQA, and NQ, with QwQ-32B as the LRM. To ensure model's general capabilities, we fine-tune GLM-4-9B \citep{DBLP:journals/corr/abs-2406-12793} with the constructed dataset (roughly 20k filtered samples), as well as the general SFT dataset from GLM-4 \citep{DBLP:journals/corr/abs-2406-12793}.

\paragraph{Evaluation.}
For MuSiQue and HotpotQA, we use the original corpora provided by the respective authors. For NQ, we follow the corpus setup from ``Lost in the middle'' \citep{DBLP:journals/tacl/LiuLHPBPL24}. To increase the difficulty, particularly for comparisons with long-context LLMs, we scale up the number of distractor documents, resulting in corpora with 48k to 58k token lengths. Thus, this design demands high-quality queries to enhance retrieval quality.

For IIRC, FRAMES, and FanOutQA, where official corpora are incomplete or unavailable, we retrieve the top 5 most relevant Wikipedia snippets via the Google Serper API. These benchmarks are excluded from the \textit{in-context} settings evaluations due to the lack of official corpora. All baselines are evaluated using their official implementations with our RAG engine and corpus setup.

\paragraph{RAG engine.} Our RAG engine consists of two main components: retrieval and generation. For retrieval, we utilize the embedding model embedding-3 from Zhipu's API\footnote{\url{https://bigmodel.cn/dev/api/vector/embedding}}, along with a reranker based on the GLM3 architecture to enhance retrieval quality. This setup is applied only to benchmarks with self-constructed corpora, while for IIRC, FRAMES, and FanOutQA, documents are retrieved directly via Google. For generation, we use GLM-4-32B with a 128k context window to generate responses based on the retrieved documents.

\subsection{Main results}
Table~\ref{tab:main-experiment} presents our main results across six benchmarks. Overall, our approach ReaRAG-9B achieves the highest average $\text{ACC}_{L}$ (48.5) across all evaluated methods. Despite the recent momentum around RL in RAG, particularly with models such as R1-Searcher and ReSearch, ReaRAG-9B trained solely with SFT to generate knowledge-guided reasoning chain, still matches or exceeds performance on MuSiQue, HotpotQA, IIRC, and NQ, and outperforms RL-based baselines on FRAMES and FanOutQA (by 5.8\%-7.3\% $\text{ACC}_{L}$), suggesting SFT can achieve equally powerful performance for open-ended QA tasks. Although designed for multi-hop tasks, ReaRAG also performs strongly on single-hop QA, achieving 55.5 $\text{ACC}_{L}$ on NQ, demonstrating ReaRAG’s capabilities beyond multi-hop reasoning. 

However, discrepancies between EM and $\text{ACC}_{L}$, e.g., on NQ, where ReaRAG achieves a higher $\text{ACC}_{L}$ than ReSearch (55.5 vs. 53.0) but yields a lower EM score (32.0 vs. 37.0), suggest that EM may fail to capture contextually valid answers, a phenomenon consistently observed across all other benchmarks, necessitating the use of $\text{ACC}_{L}$ metric.

Comparing \textit{vanilla RAG} with \textit{in-context} settings across different GLM-4 backbone scales, we find that \textit{vanilla RAG} generally performs better, suggesting that long-context models may struggle with distractor-heavy corpora. The only exception is MuSiQue with the GLM-4-32B backbone, where the \textit{in-context} setting slightly outperforms \textit{vanilla RAG} (33.50 vs. 29.00). Under \textit{vanilla RAG} setting, QwQ-32B significantly outperforms GLM-4-32B on complex multi-hop benchmarks such as FRAMES and FanOutQA, highlighting the advantage of LRMs with strong reasoning capabilities.

Notably, the original SearChain paper reports results using GPT-3.5, a proprietary large-scale model. For a fair comparison, we replace it with Qwen2.5-7B-Instruct \citep{DBLP:journals/corr/abs-2412-15115}. Under this setting, SearChain underperforms across all benchmarks. Similarly, Self-RAG, despite aiming to improve retrieval quality, lacks multi-turn retrieval strategies and shows weak performance. Search-o1, although leveraging QwQ-32B, a strong reasoning LRM to perform multi-turn retrieval, only performs on par with \textit{vanilla RAG} settings. We further analyze these findings in Section~\ref{sec:analysis}.

\subsection{Ablation}

\paragraph{Closed-book performance.} We conduct a closed-book experiment to evaluate the parametric knowledge of the language models. The results, presented in Table~\ref{tab:close-book} show that QwQ-32B outperforms GLM-4 on multi-hop benchmarks requiring strong reasoning, except for HotpotQA and FanOutQA. Nevertheless, their parametric knowledge remains insufficient compared to the results in Table~\ref{tab:main-experiment}.

\paragraph{Advantage of strong reasoning.} To evaluate the impact of strong reasoning capabilities, we fine-tune a model that lacks such abilities under the same Thought-Action-Observation paradigm. This variant, denoted as w/o reasoning in Table~\ref{tab:reasoning}, shares the same backbone architecture as ReaRAG-9B and follows the data construction process outlined in Algorithm~\ref{alg:data_construction_no_reasoning}. However, instead of leveraging a strong reasoning model like QwQ-32B for data generation, we employ GLM-4-9B, which lacks robust reasoning abilities. Unlike the previous data construction approach in Algorithm~\ref{alg:data_construction}, which used only multi-hop questions as input, we now provide GLM-4-9B with additional information, including ground-truth decompositions and ground-truth answers. The instruction prompt $\mathcal{P}_{ablat}$ used to generate its reasoning chain is detailed in Appendix~\ref{sec:prompts}.

Table~\ref{tab:reasoning} shows that ReaRAG-9B with enhanced reasoning capabilities (w/ reasoning) consistently outperforms its counterpart without reasoning, achieving a notable gain of 6-12\% $\text{ACC}_{L}$ gain on the multi-hop benchmarks and 3.5\% gain on single-hop NQ. The smaller gain on NQ suggests that strong reasoning offers limited benefit for single-hop tasks, consistent with the assumption that such questions require less compositional reasoning.


\subsection{Analysis}
\label{sec:analysis}
\begin{table}[H]
    \centering
    \setlength{\tabcolsep}{0.9pt} 
    \begin{tabular}{lcccccVc}
        \toprule
        \multirow{2}{*}{} & \multicolumn{5}{c V}{\textbf{Multi-hop}} & \textbf{Single-hop} \\
        \cmidrule(lr){2-6}\cmidrule(lr){7-7}
        & \textbf{MQ} & \textbf{HP} & \textbf{IIRC} & \textbf{FR} & \textbf{FQ} & \textbf{NQ} \\
        \midrule
        Invalid (\%) & 19.0 & 28.0 & 18.0 & 18.9 & 30.0 & 25.0 \\
        \bottomrule
    \end{tabular}
    \caption{\label{tab:invalid}
        Invalid generation rates of special tokens in QwQ-32B, leading to retrieval failures in Search-o1.}
\end{table}
\subsubsection{Performance against strong baseline}
We conduct an in-depth analysis and compare our approach against the strong baselines, including Search-o1, R1-Searcher and ReSearch. Below, we highlight key factors affecting their performance.

\paragraph{Token generation failure.} QwQ-32B struggles to follow the instruction prompt, failing to generate special tokens (e.g., $|begin\_search\_query|$) essential for Search-o1 to retrieve external knowledge. This limitation forces Search-o1 into a closed-book setting, significantly impairing its performance. Table~\ref{tab:invalid} quantifies this issue, revealing invalid generation rates of 18–30\%.

\paragraph{Information extraction failure.} Search-o1 introduces the Reason-in-Documents module, leveraging QwQ-32B for in-depth reasoning over retrieved documents to generate responses as search results. However, this module has a key limitation: it may incorrectly conclude that no useful information is available (Table~\ref{tab:casestudy_no_info}). Our analysis identifies the primary cause: the module attempts to answer the original multi-hop question based on the search query, but the retrieved information is insufficient. For example, as shown in Table~\ref{tab:casestudy_no_info}, the module searches for information related to ``Hannibal and Scipio book'' at first search, but the retrieved content only includes the book's author, lacking information about the place of education. This flaw weakens Search-o1, as it continuously searches for non-existent information, causing reasoning paths to diverge and ultimately hitting iteration limits.

\paragraph{Hallucination in Reason-in-Documents module.} The Reason-in-Documents module is prone to hallucination Table~\ref{tab:casestudy_hallucinate}. For instance, when searching for the members of Bruce Lee Band, the module fails to find relevant information and fabricates ``Less Than Records'' based on parametric knowledge rather than the provided corpus. This hallucination propagates through subsequent reasoning steps, degrading the final answer quality.

\paragraph{Poor generalization on harder benchmarks.} R1-Searcher and ReSearch show weaker performance on FRAMES and FanOutQA, despite matching ReaRAG on other benchmarks. Case studies in Tables~\ref{tab:poor_generalzation1} and~\ref{tab:poor_generalzation2} reveal that both models struggle to reason effectively in the presence of noisy information. R1-Searcher, in particular, proceeds to generate final answers despite lacking key information. These findings suggest that RL-based models lack the ability to reflect and revise earlier reasoning, an ability essential for robust multi-hop reasoning. In contrast, ReaRAG successfully realigns its reasoning path, even after early-stage errors, demonstrating greater robustness on complex multi-hop tasks.  

\begin{figure}[t]
\centering
  \includegraphics[width=1.0\linewidth]{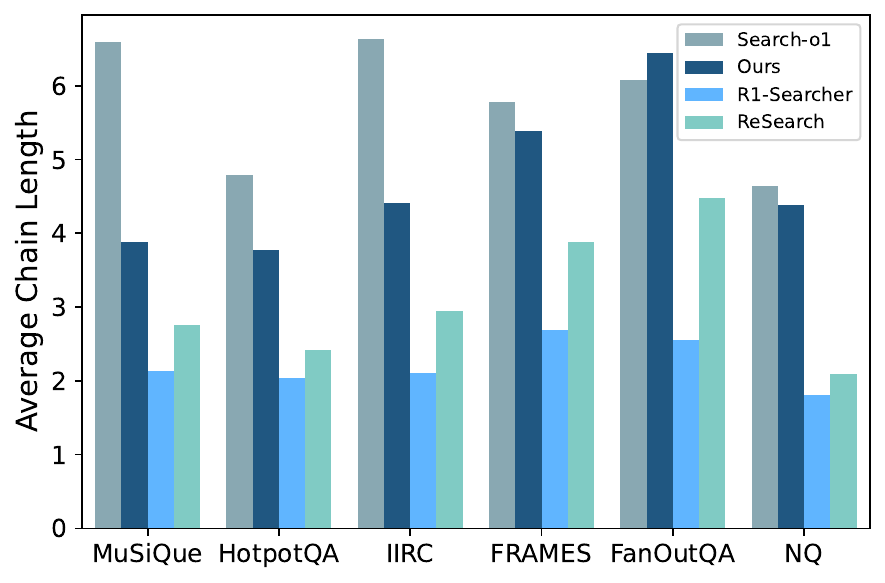}
  \caption {Comparison of chain length across all benchmarks. We measure the number of retrievals needed for the baselines to achieve a full $\text{ACC}_{L}$ score. Search-o1 consistently requires more steps than ReaRAG, highlighting the tendency of the underlying LRM to overthink in multi-hop QA tasks. }
  \label{fig:chain_length}
\end{figure}

\paragraph{Overthinking in multi-hop QA.} Recent studies have shown that LRMs tend to overthink by generating overly long reasoning chains, causing redundancy in multi-hop QA \citep{DBLP:journals/corr/abs-2412-21187, DBLP:journals/corr/abs-2501-12599}. We analyze this by comparing the number of retrievals needed to achieve full $\text{ACC}_{L}$. Compared to Search-o1, ReaRAG requires fewer steps across all benchmarks except for FanOutQA, demonstrating efficiency in multi-hop reasoning while achieving better performance. Case studies in Tables~\ref{tab:casestudy_redundant1} and~\ref{tab:casestudy_redundant2} further highlight this difference. However, for baselines such as ReSearch and R1-Searcher which are optimized for retrieval, ReaRAG consistently takes more steps. 

\subsubsection{Strength of ReaRAG}
This section showcases ReaRAG’s advanced reasoning capabilities. Table~\ref{tab:casestudy3} shows that ReaRAG initially mistakenly identified ``Anne of Austria'' as the grandmother of ``Philippe'' rather than his mother. However, ReaRAG later detected this mistake, verified the information, and corrected it. This self-correction mechanism helps prevent errors from propagating to later reasoning steps.

Table~\ref{tab:casestudy4} shows how ReaRAG resolves ambiguity in a multi-hop question via multiple retrieval. The query involves ``The Hard Easy'', which refers to both a film and a TV series. At the sixth reasoning step, ReaRAG also faces conflicting information but successfully disambiguates and provides the correct answer.

Table~\ref{tab:casestudy5} provides another example of ReaRAG handling ambiguity in a multi-hop question while resolving a knowledge conflict. Its parametric knowledge incorrectly states that Sonic is voiced by ``Roger Craig Smith'' instead of ``Jim Cummings''. ReaRAG detects and corrects this inconsistency, ultimately reaching the correct answer. This case further highlights its robust reasoning abilities.

These examples highlight ReaRAG’s ability to iteratively perform knowledge-guided reasoning. Compared to existing baselines, our approach better integrates reasoning model with external knowledge, enhancing factual accuracy.

\section{Conclusion}
In this study, we introduce \textbf{ReaRAG}, a factuality-enhanced reasoning model capable of performing knowledge-guided reasoning. ReaRAG iteratively plans and reflects on reasoning steps, leveraging external knowledge to validate each step in the reasoning chain. Through comprehensive evaluation across six benchmarks, ReaRAG achieves the best overall performance, outperforming RL-based methods on challenging benchmarks such as FRAMES and FanOutQA, while matching them on easier tasks. Unlike RL-based approaches requiring complex policy optimization and high computational cost, our method strategically distills reasoning capabilities from LRMs, offering a more efficient alternative. Further analysis shows its robustness in handling complex multi-hop questions while mitigating the overthinking behavior seen in Search-o1. These results underscore the effectiveness of strategic reasoning distillation as a practical alternative to RL-based RAG methods. 


\section*{Limitations}
\paragraph{Limited action space}
While ReaRAG demonstrates strong performance in the QA task, its action space is currently limited to only \texttt{search} and \texttt{finish} in this study. Consequently, it is restricted to processing local knowledge sources and cannot perform actions such as leveraging a code compiler for coding tasks, executing mathematical calculations, or conducting real-time web searches. Expanding its action space could enhance its adaptability across diverse problem domains.

\paragraph{Data construction efficiency}
To equip ReaRAG with a structured reasoning process, we fine-tune ReaRAG using structured responses generated by the LRM. However, this approach relies on the LRM’s strong instruction-following ability, and a substantial portion of the data is discarded due to validity issues, leading to computational inefficiency and resource waste. Improving data augmentation techniques could mitigate this limitation.

\paragraph{Inference latency}  
ReaRAG solves questions iteratively, requiring multiple reasoning steps to reach the final answer. While this enhances accuracy, it also increases inference time compared to models that generate answers in a single pass. This trade-off between reasoning depth and efficiency may limit its practicality in real-time applications or scenarios with strict latency constraints.

\section*{Acknowledgments}
This work is supported by National Natural Science Foundation of China (62476150), Beijing Natural Science Foundation (L243006), Tsinghua University Initiative Scientific Research Program and Tsinghua University (Department of Computer Science and Technology)-Siemens Ltd., China Joint Research Center for Industrial Intelligence and Internet of Things (JCIIOT).

This paper involved the use of AI-assisted tools (e.g., ChatGPT) for language refinement and editing. All content was reviewed and verified by the authors.

\bibliography{main}



\newpage
\appendix
\onecolumn

\section*{Appendix}

\section{Ablation}


\begin{table}[H]
    \centering
    \resizebox{1\linewidth}{!}{
    \begin{tabular}{l *{10}{c} V *{2}{c} V c}
    \toprule
    \multirow{3}{*}{\textbf{Model}} 
    & \multicolumn{10}{c V}{\textbf{Multi-hop}}
    & \multicolumn{2}{c V}{\textbf{Single-hop}}
    & \multirow{3}{*}{\textbf{Average}} \\
    \cmidrule(r){2-11}\cmidrule(l){12-13}
     & \multicolumn{2}{c}{\textbf{MuSiQue}} 
     & \multicolumn{2}{c}{\textbf{HotpotQA}} 
     & \multicolumn{2}{c}{\textbf{IIRC}} 
     & \multicolumn{2}{c}{\textbf{FRAMES}} 
     & \multicolumn{2}{c V}{\textbf{FanOutQA}}
     & \multicolumn{2}{c V}{\textbf{NQ}} \\
    \cmidrule(lr){2-3}\cmidrule(lr){4-5}\cmidrule(lr){6-7}\cmidrule(lr){8-9}\cmidrule(lr){10-11}\cmidrule(lr){12-13}\cmidrule(l){14-14}
     & $\text{ACC}_{L}$ & EM 
     & $\text{ACC}_{L}$ & EM 
     & $\text{ACC}_{L}$ & EM 
     & $\text{ACC}_{L}$ & EM 
     & $\text{ACC}_{L}$ & R-L 
     & $\text{ACC}_{L}$ & EM
     & $\text{ACC}_{L}$ \\
    \midrule
    GLM-4-9B\small{(128k)} 
            & 3.5 & 0.0
            & 29.5 & 23.0
            & 17.0 & \textbf{15.0}
            & 5.2 & 3.5 
            & 7.1 & 18.0
            & 27.50 & 16.00
            & 15.0 \\
        GLM-4-32B\small{(128k)} 
            & 6.5 & 1.0
            & \textbf{40.0} & \textbf{28.0}
            & 17.0 & 11.5
            & 10.5 & 4.6 
            & \textbf{11.9} & \textbf{22.5} 
            & \textbf{44.5} & \textbf{25.0}
            & \textbf{21.7} \\
        QwQ-32B 
            & \textbf{11.0} & \textbf{2.0}
            & 35.0 & 10.0
            & \textbf{21.3} & 14.5
            & \textbf{14.4} & \textbf{9.1} 
            & 11.0 & 13.2 
            & 37.5 & 12.0 
            & \textbf{21.7} \\ 
    \bottomrule
    \end{tabular}
    }
    \caption{\label{tab:close-book}
        Closed-book performance of language models on multi-hop and single-hop benchmarks. These models perform better on single-hop benchmarks but score significantly lower on multi-hop benchmarks, highlighting the limitations of relying solely on parametric knowledge for these benchmarks.}
\end{table}

\begin{table}[H]
    \centering
    \resizebox{1\linewidth}{!}{
    \begin{tabular}{l *{10}{c} V *{2}{c} V c}
    \toprule
    \multirow{3}{*}{\textbf{Model}} 
    & \multicolumn{10}{c V}{\textbf{Multi-hop}}
    & \multicolumn{2}{c V}{\textbf{Single-hop}}
    & \multirow{3}{*}{\textbf{Average}} \\
    \cmidrule(r){2-11}\cmidrule(l){12-13}
     & \multicolumn{2}{c}{\textbf{MuSiQue}} 
     & \multicolumn{2}{c}{\textbf{HotpotQA}} 
     & \multicolumn{2}{c}{\textbf{IIRC}} 
     & \multicolumn{2}{c}{\textbf{FRAMES}} 
     & \multicolumn{2}{c V}{\textbf{FanOutQA}}
     & \multicolumn{2}{c V}{\textbf{NQ}} \\
    \cmidrule(lr){2-3}\cmidrule(lr){4-5}\cmidrule(lr){6-7}\cmidrule(lr){8-9}\cmidrule(lr){10-11}\cmidrule(lr){12-13}\cmidrule(l){14-14}
     & $\text{ACC}_{L}$ & EM 
     & $\text{ACC}_{L}$ & EM 
     & $\text{ACC}_{L}$ & EM 
     & $\text{ACC}_{L}$ & EM 
     & $\text{ACC}_{L}$ & R-L 
     & $\text{ACC}_{L}$ & EM
     & $\text{ACC}_{L}$ \\
    \midrule
    \textbf{ReaRAG-9B} 
        & & & & & & & & \\[-2pt]
        \quad \small{- \textit{w/o reasoning}}
          & 57.5 & 38.0
          & 69.5 & 50.0
          & 31.8 & 22.0
          & 26.6 & 19.2
          & 5.5 & 21.6
          & 52.0 & 29.0
          & 40.5 \\
        \quad \small{- \textit{w/ reasoning}}
          & \textbf{67.0} & \textbf{46.0}
          & \textbf{81.5} & \textbf{58.0}
          & \textbf{41.0} & \textbf{27.5}
          & \textbf{34.2} & \textbf{21.2}
          & \textbf{11.6} & \textbf{24.3}
          & \textbf{55.5} & \textbf{32.0}
          & \textbf{48.5}\\
    \bottomrule
    \end{tabular}
    }
    \caption{\label{tab:reasoning}
        Performance comparison of models with and without strong reasoning capabilities. \textit{w/ reasoning} consistently outperforms \textit{w/o reasoning} across all benchmarks, demonstrating the effectiveness of our fine-tuning process, which enables ReaRAG-9B to inherit the strong reasoning abilities of LRM.}
\end{table}

\begin{algorithm}
\caption{Data Construction to fine-tune ReaRAG w/o reasoning}
\label{alg:data_construction_no_reasoning}
\textbf{Input:} Seed Dataset $\mathcal{D_{\text{seed}}}$, Large Language Model $\mathcal{M}_{LLM}$, Instruction Prompt $\mathcal{P}_{ablat}$ \\
\textbf{Output:} Dataset $\mathcal{D}_{\text{ablat}}$
\begin{algorithmic}[1]
\State Initialize $\mathcal{D}_{\text{ablat}} \gets \emptyset$
\For{each $(x_i, y_i, decomp_i) \sim \mathcal{D_{\text{seed}}}$}
    \Statex \Comment{Sample question $x_i$, ground truth answer $y_i$ and golden decomposition $decomp_i$}
    \State $r'_i \gets \mathcal{M}_{LLM}([\mathcal{P}_{ablat} \oplus x_i \oplus y_i \oplus decomp_i])$
    \Statex \Comment{Generate response}
    \State $\mathcal{C}_i = [\{\tau_t, \alpha_t, o_t\}_{t=1}^N] \gets \texttt{parse}(r'_i)$ 
    \Statex \Comment{Parse a list of thought $\tau_t$, action $\alpha_t$ and observation $o_t$ into reasoning chain $\mathcal{C}_i$}
    \State Append $\mathcal{C}_i$ to $\mathcal{D}_{\text{ablat}}$
\EndFor
\State \textbf{return} $\mathcal{D}_{\text{ablat}}$
\end{algorithmic}
\end{algorithm}

\section{Prompts}
\label{sec:prompts}

\begin{figure}[H]
\begin{tcolorbox}[colback=gray!10, colframe=darkgray, title=Instruction prompts $\mathcal{P}_d$ for data construction to fine-tune ReaRAG, width=\textwidth]
\small
Your task is to solve a question answering task. To improve your solving accuracy, please conduct reasoning processes following this sequence: Thought, Action, Observation steps. Thought can reason about the current situation, and Action is in the form of a function. There are two available function types:

\vspace{5pt}
\textbf{Available Functions:}

(1) Search
{\footnotesize
\begin{lstlisting}[basicstyle=\ttfamily\footnotesize, breaklines=true, columns=flexible]
{
    "name": "search",
    "description": "It can help you find useful information through the internet or local knowledge base. You can use this tool to access external knowledge.",
    "parameters": {
        "type": "object",
        "properties": {
            "query": {
                "description": "what you want to search"
            }
        },
        "required": ["query"]
    }
}
\end{lstlisting}
}

(2) Finish
{\footnotesize
\begin{lstlisting}[basicstyle=\ttfamily\footnotesize, breaklines=true, columns=flexible]
{
    "name": "finish",
    "description": "You can use this function to make a conclusion from the reasoning process and give the final answer. The reasoning process is completed after this `finish` function is called",
    "parameters": {
        "type": "object",
        "properties": {
            "answer": {
                "description": "the final answer"
            }
        },
        "required": ["answer"]
    }
}
\end{lstlisting}
}

\vspace{5pt}
\textbf{Some important rules you must follow:} \\
(1) Please follow the function calling format above strictly. \\
(2) A set of `Thought`, `Action`, and `Observation` is considered as one reasoning step. Add numbering after each `Thought`, `Action`, and `Observation` to indicate the sequence of the reasoning steps. \\
(3) Please give your `Thought` first, then the `Action`, and finally the `Observation`, follow the format as shown in the in-context examples below. \\
(4) In your `Thought`, you should perform reflection when necessary to ensure the correctness of your reasoning process, such as: "Wait! Maybe I made some mistakes! I need to rethink from scratch", "Alternatively, we can...", "Hold on, let's try another approach", etc. \\
(5) Give your `Action` in the form of function call, as shown in in-context examples below. \\
(6) You should not provide information based on your own knowledge, only use the information provided in the context.

\vspace{5pt}
\textbf{Some example of reflection text:} \\
"There is no enough information from the previous steps. I need to plan my query again." \\
"No context found from observation. Let me restart the reasoning process." \\
"Missing information. Let me restructure my query." \\
"Wait! Maybe I made some mistakes! I need to rethink from scratch." \\
"I think I need to take a step back and reconsider my approach." \\
"I need to reevaluate my reasoning process. Let's start over." \\
"I need to reflect on my reasoning. Let's try a different approach." \\
\textcolor{blue}{\{More examples of reflection text. Simplfied for readability}\} \\

\vspace{5pt}
\textbf{In-Context Example:} \\
\textcolor{blue}{\{Some in-context examples}\}

\end{tcolorbox}
\end{figure}

\begin{figure}[H]
\begin{tcolorbox}[colback=gray!10, colframe=darkgray, title=Instruction prompt $\mathcal{P}$ for fine-tuning and inference with ReaRAG, width=\textwidth]
\small

Your task is to solve a question answering task. To improve your solving accuracy, please conduct reasoning process interleaving Thought, Action, Observation steps. Thought can reason about the current situation, and Action are in the form of function, there are two types:

\vspace{5pt}
\textbf{Available Functions:}

(1) Search
{\footnotesize
\begin{lstlisting}[basicstyle=\ttfamily\footnotesize, breaklines=true, columns=flexible]
{
    "name": "search",
    "description": "It can help you find useful information through the internet or local knowledge base. You can use this tool to access external knowledge.",
    "parameters": {
        "type": "object",
        "properties": {
            "query": {
                "description": "what you want to search"
            }
        },
        "required": ["query"]
    }
}
\end{lstlisting}
}

(2) Finish
{\footnotesize
\begin{lstlisting}[basicstyle=\ttfamily\footnotesize, breaklines=true, columns=flexible]
{
    "name": "finish",
    "description": "You can use this function to make a conclusion from the reasoning process and give the final answer. The reasoning process is completed after this `finish` function is called",
    "parameters": {
        "type": "object",
        "properties": {
            "answer": {
                "description": "the final answer"
            }
        },
        "required": ["answer"]
    }
}
\end{lstlisting}
}

\vspace{5pt}

Please follow the format strictly.

\end{tcolorbox}
\end{figure}

\newpage

\begin{figure}[H]
\begin{tcolorbox}[halign=flush left, colback=gray!10, colframe=darkgray, title=Answer prompt $\mathcal{P}_{ans}$ to derive final answer, width=\textwidth]
\small
<|system|> \\
You are a QA assistant. Always return a short answer. Output ONLY the answer with no extra words.<|user|> \\
You have conducted multiple searches to seek for informations to answer the Question. \\
\textbf{Searches} \\
\textcolor{blue}{\{Previous search query and search results}\} \\

\vspace{6pt}

\textbf{Instruction} \\
Answer the Question based on the searches. **Only output the final answer. Do NOT add any explanation, punctuation, or extra words.** \\
Question: \textcolor{blue}{\{question}\} \\  
Answer: <|assistant|>

\vspace{3pt}
\end{tcolorbox}
\end{figure}

\begin{figure}[H]
\begin{tcolorbox}[halign=flush left, colback=gray!10, colframe=darkgray, title=Judgement prompt $\mathcal{P}_{judge}$ for $\text{ACC}_{L}$ metrics on all benchmarks except for FanOutQA , width=\textwidth]
\small
You are asked to evaluate the quality of the AI assistant's answer to user questions as an impartial judge, and your evaluation should take into account factors including correctness (high priority), and comprehensiveness (whether the assistant's answer covers all points). \\
Read the AI assistant's answer and compare against the reference answer, and give an overall integer rating in 1, 2, 3 (1 = wrong or irrelevant, 2 = partially correct, 3 = correct and comprehensive) based on the above principles, strictly in the following format:"[[rating]]", e.g. "[[2]]". \\

\vspace{6pt}

\text{[Reference answer]} \\
\textcolor{blue}{\{Ground truth}\} \\ 
\text{[Assistant's answer]} \\
\textcolor{blue}{\{Prediction}\} \\ 
Rating: \\

\vspace{3pt}
\end{tcolorbox}
\end{figure}

\begin{figure}[H]
\begin{tcolorbox}[halign=flush left, colback=gray!10, colframe=darkgray, title=Judgement prompt $\mathcal{P}_{judge}$ for $\text{ACC}_{L}$ metrics on FanOutQA benchmarks, width=\textwidth]
\small
\text{[BEGIN DATA]} \\
************ \\
\text{[Question]}: \textcolor{blue}{\{Question}\} \\
************ \\
\text{[Expert]}: \textcolor{blue}{\{Ground truth}\} \\
************ \\
\text{[Submission]}: \textcolor{blue}{\{Prediction}\} \\
************ \\
\text{[END DATA]} \\

\vspace{8pt}

Compare the factual content of the submitted answer with the expert answer. Ignore any differences in style, grammar, or punctuation. \\
The submitted answer may either be a subset or superset of the expert answer, or it may conflict with it. Determine which case applies. First, write out your reasoning in a step-by-step manner to carefully evaluate the factual content. Make sure your conclusion is supported by detailed reasoning. Avoid simply stating the correct choice immediately.  \\
After completing your reasoning, output only the final selected letter (A, B, C, D, E, or F). Format your output exactly as follows: wrap the letter inside <answer> and </answer> tags. Important: Do not include any extra words, quotation marks, punctuation, or explanations — only the letter inside the tags. \\

\vspace{8pt}

Here are the available choices:\\
(A) The submitted answer is a subset of the expert answer and is fully consistent with it.\\
(B) The submitted answer is a superset of the expert answer and is fully consistent with it.\\
(C) The submitted answer contains all the same details as the expert answer.\\
(D) There is a disagreement between the submitted answer and the expert answer.\\
(E) The answers differ , but these differences don't matter from the perspective of factuality.\\
(F) The submitted answer does not answer the question or is otherwise invalid.\\

\vspace{3pt}
\end{tcolorbox}
\end{figure}

\begin{figure}[H]
\begin{tcolorbox}[colback=gray!10, colframe=darkgray, title=Instruction prompts $\mathcal{P}_{ablat}$ for data construction to fine-tune ReaRAG w/o reasoning, width=\textwidth]
\small
You are given Question, Ground-truth answer, and Decompositions, your task is to give reasoning process interleaving Thought, Action, Observation steps. Thought can reason about the current situation, and Action are in the form of function, there are two types:

\vspace{5pt}
(1) search
{\footnotesize
\begin{lstlisting}[basicstyle=\ttfamily\footnotesize, breaklines=true, columns=flexible]
{
    "name": "search",
    "description": "It can help you find useful information through the internet or local knowledge base. You can use this tool to access external knowledge.",
    "parameters": {
        "type": "object",
        "properties": {
            "query": {
                "description": "what you want to search"
            }
        },
        "required": ["query"]
    }
}
\end{lstlisting}
}

(2) finish
{\footnotesize
\begin{lstlisting}[basicstyle=\ttfamily\footnotesize, breaklines=true, columns=flexible]
{
    "name": "finish",
    "description": "You can use this function to make a conclusion from the reasoning process and give the final answer. The reasoning process is completed after this `finish` function is called",
    "parameters": {
        "type": "object",
        "properties": {
            "answer": {
                "description": "the final answer"
            }
        },
        "required": ["answer"]
    }
}
\end{lstlisting}
}

\vspace{5pt}
After `finish` function being called, the reasoning process is completed, then marked with <|completed|>. Please follow the format strictly. \\

\vspace{5pt}

Here are in-context examples: \\
\textbf{Question}: \\
What was the wettest year in the second largest city in the state where Yuma's Library District is located? \\

\vspace{5pt}

\textbf{Ground-truth answer}:\\
1905 \\

\vspace{5pt}

\textbf{Decompositions}:
{\footnotesize
\begin{lstlisting}[basicstyle=\ttfamily\footnotesize, breaklines=true, columns=flexible]
[
    {"question": "Yuma located in the administrative territorial entity",
        "answer": "Yuma County",
        "context": "Some context" # Simplified for readability},
    ... # Simplified for readability
    {"question": "Tucson's wettest year?",
        "answer": "1905",
        "context": "Some context" # Simplified for readability}
]
\end{lstlisting}
}

Reasoning process with function call: \\
\textcolor{blue}{\{Reasoning process example}\}

\end{tcolorbox}
\end{figure}


\section{Case study}
\begin{table}[H]
    \centering
    \small
    \renewcommand{\arraystretch}{1.3}
    \begin{tabular}{|p{0.95\linewidth}|}
        \hline
        \rowcolor{lightgray}\textbf{\# Case Study 1: Where was the author of Hannibal and Scipio educated at? } \\
        \hline
        \textbf{Ground truth:} Exeter College \\
        \hline
        \rowcolor{lightgray}\textbf{Search-o1} \\
        \hline
        Alright, I've got this question: \colorbox{lightgray}{\textit{... Simplified here.}}\\
        So, I'll do a search for "Hannibal and Scipio book."\\\textcolor{teal}{\texttt{<|begin\_search\_query|>}}Hannibal and Scipio book\textcolor{teal}{\texttt{<|end\_search\_query|>}}\\\textcolor{blue}{\texttt{<|begin\_search\_result|>}}\textbf{No helpful information found.}\textcolor{blue}{\texttt{<|end\_search\_result|>}}\\Hmm, that didn't turn up anything. Maybe the book has a different title. Let me think differently. \\
        \colorbox{lightgray}{\textit{... Simplified here. Search-o1 reasoning diverges and hence unable to find related information.}}\\
        \colorbox{lightgray}{\textit{Only search queries and results are listed below for simplicity.}}
        \\\textcolor{teal}{\texttt{<|begin\_search\_query|>}}Punic Wars book\textcolor{teal}{\texttt{<|end\_search\_query|>}}\\\textcolor{blue}{\texttt{<|begin\_search\_result|>}}No helpful information found.\textcolor{blue}{\texttt{<|end\_search\_result|>}}\\\textcolor{teal}{\texttt{<|begin\_search\_query|>}}Hannibal vs. Scipio book\textcolor{teal}{\texttt{<|end\_search\_query|>}}\\\textcolor{blue}{\texttt{<|begin\_search\_result|>}}No helpful information found.\textcolor{blue}{\texttt{<|end\_search\_result|>}}\\
        \textcolor{teal}{\texttt{<|begin\_search\_query|>}}Adrian Goldsworthy books\textcolor{teal}{\texttt{<|end\_search\_query|>}}\\\textcolor{blue}{\texttt{<|begin\_search\_result|>}}No helpful information found.**Thoughts**After carefully reviewing the three web pages provided, it is evident that none of them \colorbox{lightgray}{\textit{... Simplified here.}}\textcolor{blue}{\texttt{<|end\_search\_result|>}}\\
        \textcolor{teal}{\texttt{<|begin\_search\_query|>}}Robert F. Pennell education\textcolor{teal}{\texttt{<|end\_search\_query|>}}\\\textcolor{blue}{\texttt{<|begin\_search\_result|>}}No helpful information found.\textcolor{blue}{\texttt{<|end\_search\_result|>}}\\
        \colorbox{lightgray}{\textit{... Simplified here. Search-o1 tries many unrelated search queries, demonstrating divergent reasoning.}}\\
        I've reached the maximum number of searches. Based on the information I have, I need to make an educated guess.\colorbox{lightgray}{\textit{... Simplified here. Maximum searches reaches}}\\
        \colorbox{lightgray}{\textit{Search-o1 tries to make a guess when there is not enough information to answer the multi-hop question.}} \\
        Considering that Oxford and Cambridge are traditional centers for classical studies in the UK, I'll go with Oxford.\\Therefore, my answer is:\\
        \fbox{Oxford}
        
        \\
        \hline
        \rowcolor{lightgray}\textbf{ReaRAG-9B} \\
        \hline

        \textbf{Thought 1:} So I need to find out where the author of "Hannibal" and "Scipio" was educated. First, I need to know who the author is. \\
        \textbf{Action 1:} \textcolor{pygreen}{\{\textquotesingle function\textquotesingle : \textquotesingle search\textquotesingle , \textquotesingle parameters\textquotesingle : \{\textquotesingle query\textquotesingle : \textquotesingle Who is the author of Hannibal and Scipio?\textquotesingle \}\}} \\
        \textbf{Observation 1:} The author of "Hannibal and Scipio" is Thomas Nabbes. \\
        \cline{1-1}
        
        \textbf{Thought 2:} Okay, now that I know the author is Thomas Nabbes, I need to find out where he was educated. \\
        \textbf{Action 2:} \textcolor{pygreen}{\{\textquotesingle function\textquotesingle : \textquotesingle search\textquotesingle , \textquotesingle parameters\textquotesingle : \{\textquotesingle query\textquotesingle : \textquotesingle Where was Thomas Nabbes educated?\textquotesingle \}\}} \\
        \textbf{Observation 2:} Thomas Nabbes was educated at Exeter College, Oxford. \\
        \cline{1-1}
        
        \textbf{Thought 3:} Great, so Thomas Nabbes studied at Exeter College in Oxford. I should make sure that this the correct information. \\
        \textbf{Action 3:} \textcolor{pygreen}{\{\textquotesingle function\textquotesingle : \textquotesingle search\textquotesingle , \textquotesingle parameters\textquotesingle : \{\textquotesingle query\textquotesingle : \textquotesingle Confirm the educational background of Thomas Nabbes\textquotesingle \}\}} \\
        \textbf{Observation 3:} Thomas Nabbes was educated at Exeter College, Oxford in 1621. He left the university without taking a degree. \\
        \cline{1-1}
        
        \textbf{Thought 4:} So, he studied at Exeter College in Oxford, but didn't finish his degree. That's interesting. I think this is the information I need.\\
        \textbf{Action 4:} \textcolor{pygreen}{\{\textquotesingle function\textquotesingle : \textquotesingle finish\textquotesingle , \textquotesingle parameters\textquotesingle : \{\textquotesingle answer\textquotesingle : \textquotesingle Thomas Nabbes was educated at Exeter College, Oxford.\textquotesingle \}\}} \\

        \cline{1-1}
        \textbf{Final Answer:} Exeter College, Oxford \\
        \hline
    \end{tabular}
    \caption{We compare the generations of Search-o1 and ReaRAG. During the initial searches, the Reason-in-Documents module of Search-o1 returns ``\textbf{No helpful information found}'' because it attempts to answer the multi-hop question using information retrieved based on the search query rather than the original question. Hence, the retrieved content lacks the full answer, misleading Search-o1’s reasoning. Ultimately, it reaches its search limits and fails to provide the correct answer. In contrast, ReaRAG demonstrates logical reasoning and correctly answers the query.}
    \label{tab:casestudy_no_info}
\end{table}

\begin{table}[H]
    \centering
    \small
    \renewcommand{\arraystretch}{1.3}
    \begin{tabular}{|p{0.95\linewidth}|}
        \hline
        \rowcolor{lightgray}\textbf{\# Case Study 2: What record label did the person who is part of The Bruce Lee Band start? } \\
        \hline
        \textbf{Ground truth:} Asian Man Records \\
        \hline
        \rowcolor{lightgray}\textbf{Search-o1} \\
        \hline
        Alright, so I need to find out what record label the person who is part of The Bruce Lee Band \colorbox{lightgray}{\textit{... Simplified here.}}\\ 
        So, I'll do a search for "The Bruce Lee Band."\\\textcolor{teal}{\texttt{<|begin\_search\_query|>}}The Bruce Lee Band\textcolor{teal}{\texttt{<|end\_search\_query|>}}\\\textcolor{blue}{\texttt{<|begin\_search\_result|>}}\textbf{No helpful information found.}\textcolor{blue}{\texttt{<|end\_search\_result|>}}\\Hmm, that's strange. Maybe the band name is different or not well-known. Perhaps I should try searching for "Bruce Lee tribute band" or something like that.\\
        \colorbox{lightgray}{\textit{... Simplified here. Search-o1 tries different searches, but Reason-in-Documents returns ``No helpful information''}}\\
        \colorbox{lightgray}{\textit{Only search queries and search results are listed here for simplicity}}
        \\\textcolor{teal}{\texttt{<|begin\_search\_query|>}}Bruce Lee tribute band\textcolor{teal}{\texttt{<|end\_search\_query|>}}\\\textcolor{blue}{\texttt{<|begin\_search\_result|>}}\textbf{No helpful information found.}\textcolor{blue}{\texttt{<|end\_search\_result|>}}\\
        \textcolor{teal}{\texttt{<|begin\_search\_query|>}}Bruce Lee band\textcolor{teal}{\texttt{<|end\_search\_query|>}}\\\textcolor{blue}{\texttt{<|begin\_search\_result|>}}\textbf{No helpful information found.}\textcolor{blue}{\texttt{<|end\_search\_result|>}}\\
        \colorbox{lightgray}{\textit{... Simplified here.}}\\
        \textcolor{teal}{\texttt{<|begin\_search\_query|>}}The Bruce Lee Band members\textcolor{teal}{\texttt{<|end\_search\_query|>}}\\\textcolor{blue}{\texttt{<|begin\_search\_result|>}}\textcolor{red}{Less Than Records}, started by a member of Less Than Jake, who is part of The Bruce Lee Band.**Explanation**Based on the information from Web Page 1, The Bruce Lee Band features Less Than Jake as part of its backing band. Less Than Jake is known for starting their own record label, \textcolor{red}{Less Than Records.} \colorbox{lightgray}{\textit{... Simplified here. Search-o1 fabricates the name "Less Than Records"}}\\
        \textcolor{teal}{\texttt{<|begin\_search\_query|>}}Less Than Jake record label\textcolor{teal}{\texttt{<|end\_search\_query|>}}\\\textcolor{blue}{\texttt{<|begin\_search\_result|>}}No helpful information found.**Explanation**After reviewing the three web pages provided for the search query "Less Than Jake record label," none of them contained specific information about which member of Less Than Jake started \textcolor{red}{Less Than Records}. \colorbox{lightgray}{\textit{... Simplified here.}}\textcolor{blue}{\texttt{<|end\_search\_result|>}}\\
        \colorbox{lightgray}{\textit{... Simplified here.}}\textcolor{red}{Less Than Records} was started by a member of Less Than Jake, who is also part of The Bruce Lee Band.Therefore, the record label started by the person who is part of The Bruce Lee Band is \textcolor{red}{Less Than Records}.\\\textbf{Final Answer}\\
        \fbox{Less Than Records}
        \\
        \hline
        \rowcolor{lightgray}\textbf{ReaRAG-9B} \\
        \hline

        \textbf{Thought 1:} I need to find out which record label the person who is part of The Bruce Lee Band started. First, I should figure out who is part of The Bruce Lee Band. \\
        \textbf{Action 1:} \textcolor{pygreen}{\{\textquotesingle function\textquotesingle : \textquotesingle search\textquotesingle , \textquotesingle parameters\textquotesingle : \{\textquotesingle query\textquotesingle : \textquotesingle Who is part of The Bruce Lee Band?\textquotesingle \}\}} \\
        \textbf{Observation 1:} Mike Park and his backing band, which has included Less Than Jake and the Rx Bandits, are part of The Bruce Lee Band. \\
        \cline{1-1}
        
        \textbf{Thought 2:} So, Mike Park is involved with The Bruce Lee Band. I need to find out if he started a record label. \\
        \textbf{Action 2:} \textcolor{pygreen}{\{\textquotesingle function\textquotesingle : \textquotesingle search\textquotesingle , \textquotesingle parameters\textquotesingle : \{\textquotesingle query\textquotesingle : \textquotesingle Did Mike Park start a record label?\textquotesingle \}\}} \\
        \textbf{Observation 2:} Yes, Mike Park started a record label. \\
        \cline{1-1}
        
        \textbf{Thought 3:} Great, now I need to find out the name of the record label he started. \\
        \textbf{Action 3:} \textcolor{pygreen}{\{\textquotesingle function\textquotesingle : \textquotesingle search\textquotesingle , \textquotesingle parameters\textquotesingle : \{\textquotesingle query\textquotesingle : \textquotesingle What record label did Mike Park start?\textquotesingle \}\}} \\
        \textbf{Observation 3:} Mike Park started the record label Asian Man Records. \\
        \cline{1-1}

        \textbf{Thought 4:} So, Asian Man Records is the label Mike Park started. But I should make sure that this is correct and see if there's any other relevant information. \\
        \textbf{Action 4:} \textcolor{pygreen}{\{\textquotesingle function\textquotesingle : \textquotesingle search\textquotesingle , \textquotesingle parameters\textquotesingle : \{\textquotesingle query\textquotesingle : \textquotesingle Confirm if Asian Man Records was started by Mike Park.\textquotesingle \}\}} \\
        \textbf{Observation 4:} Yes, Asian Man Records was started by Mike Park. \\
        \cline{1-1}
        
        \textbf{Thought 5:} Alright, I have confirmed that Mike Park started Asian Man Records. Since he is part of The Bruce Lee Band, this should be the answer. \\
        \textbf{Action 5:} \textcolor{pygreen}{\{\textquotesingle function\textquotesingle : \textquotesingle finish\textquotesingle , \textquotesingle parameters\textquotesingle : \{\textquotesingle answer\textquotesingle : \textquotesingle Asian Man Records\textquotesingle \}\}} \\

        \cline{1-1}
        \textbf{Final Answer:} Asian Man Records \\
        \hline
    \end{tabular}
    \caption{We compare the generations of Search-o1 and ReaRAG. Search-o1's Reason-in-Documents module fabricates non-existent information ("\textcolor{red}{Less Than Records}", highlighted in \textcolor{red}{red}), compromising factual accuracy and leading to a hallucinated response. In contrast, ReaRAG successfully conduct knowledge-guided reasoning to generate a factual answer.}
    \label{tab:casestudy_hallucinate}
\end{table}

\renewcommand{\arraystretch}{1.3}         
\setlength{\tabcolsep}{3pt}               

\begin{center}  
    \small
    \begin{longtable}{|p{0.95\linewidth}|}
        \hline
        \rowcolor{lightgray}\textbf{\# Case Study 3: What is the scientific name of an amphibian that is listed as endangered by the Canadian government (as of 2024), and its only population in Canada occurs on an island which is the southernmost inhabited part of Canada?}\\
        \hline
        \endfirsthead
        
        \multicolumn{1}{l}{\small \emph{Table~\ref{tab:poor_generalzation1} –
continued from previous page}}\\[0.3em]
        \hline
        \rowcolor{lightgray}
        \multicolumn{1}{|p{0.95\linewidth}|}{\textbf{\# Case Study 3 (continued)}} \\
        \hline
        \endhead
        
        \hline
        \multicolumn{1}{r}{\small \emph{Continued on next page}}\\[-0.6em]
        \endfoot
        
        \endlastfoot
        
        \textbf{Ground truth:} Ambystoma texanum\\
        \hline
        \rowcolor{lightgray}\textbf{R1-Searcher} \\
        \hline
        To solve this question, I need to find the scientific name of an amphibian that is listed as endangered by the Canadian government and has its only population in Canada on an island that is the southernmost inhabited part of Canada. \\

        Step 1: Identify the island that is the southernmost inhabited part of Canada. \\
        Step 2: Determine which amphibian is listed as endangered by the Canadian government and has its only population on that island. \\
        
        **Step 1: Identify the island that is the southernmost inhabited part of Canada.** \\
        
        I'll use a query to find out: \\
        
        \textcolor{teal}{\texttt{<|begin\_of\_query|>}} southernmost inhabited part of Canada\textcolor{teal}{\texttt{<|end\_of\_query|>}} \\
        
        \textcolor{blue}{\texttt{<|begin\_of\_documents|>}} \\
        \textbf{Pelee, Ontario:} \\
        Pelee is the southernmost inhabited part of Canada and is situated on two major migratory bird routes, the Atlantic Flyway and the Mississippi flyway. The \\
        
        \textbf{Extreme points of Canada:}\\
        The southernmost point is Middle Island, in Lake Erie, Ontario (41°41'N 82°40'W); the southernmost water point lies just south of the island, on the Ontario– \\
        
        \textbf{Middle Island (Lake Erie):} \\
        Middle Island is a small island, just 18.5 hectares (46 acres) in area. [1] It is the southernmost point of land in Canada, located at 41°41'N, 82°41"W. \\
        
        \textbf{Extreme points of Canadian provinces:} \\
        British Columbia, Yukon and Northwest Territories border \colorbox{lightgray}{\textit{... Simplified here.}} \\
        
        \textbf{Southern Ontario:} \\
        Southern Ontario is a primary region of the Canadian province of Ontario. \colorbox{lightgray}{\textit{... Simplified here.}}\textcolor{blue}{\texttt{<|end\_of\_documents|>}} \\

        \colorbox{lightgray}{\textit{The keyword "southernmost" appears in multiple documents, but only Pelee, Ontario is correct as the southernmost}}\\
        \colorbox{lightgray}{\textit{inhabited part of Canada. R1-Searcher fails to identify it accurately.}}\\
        From the search results, I can see that \textcolor{red}{Middle Island} is the southernmost inhabited part of Canada. \\
        
        **Step 2: Determine which amphibian is listed as endangered by the Canadian government and has its only population on Middle Island.** \\
        
        I'll use a query to find out: \\
        \textcolor{teal}{\texttt{<|begin\_of\_query|>}} amphibian listed as endangered by the Canadian government with its only population on Middle Island, Ontario\textcolor{teal}{\texttt{<|end\_of\_query|>}} \colorbox{lightgray}{\textit{Wrong query due to earlier mistakes. Middle Island is not the correct place.}}\\
        
        \textcolor{blue}{\texttt{<|begin\_of\_documents|>}} \\
        \textbf{List of amphibians of Canada:} \\
        List of amphibians of Canada \colorbox{lightgray}{\textit{... Simplified here.}} \\
        
        \textbf{List of Wildlife Species at Risk (Canada):} \\
        As of November 2024, \colorbox{lightgray}{\textit{... Simplified here.}} \\
        
        \textbf{List of endangered amphibians:} \\
        Of all evaluated amphibian species, \colorbox{lightgray}{\textit{... Simplified here.}} \\
        
        \textbf{[PDF] Midland Painted Turtle - Species at risk public registry:} \\
        Painted Turtle is recognized as a species of Least Concern \colorbox{lightgray}{\textit{... Simplified here.}} \\
        
        \textbf{American burying beetle (Nicrophorus americanus): COSEWIC :} \\
        The American Burying Beetle \colorbox{lightgray}{\textit{... Simplified here.}}.\textcolor{blue}{\texttt{<|end\_of\_documents|>}} \\
        
        From the search results, I can see that there is no specific amphibian listed as endangered by the Canadian government with its only population on Middle Island, Ontario. However, I will focus on amphibians that are endangered and have populations in Canada.\textcolor{brown}{\texttt{</think>}} \\
        
         <answer> Midland Painted Turtle </answer> \\
        \hline
        \hline
        \rowcolor{lightgray}\textbf{ReSearch} \\
        \hline
        \textcolor{brown}{\texttt{<think>}} To find the scientific name of the amphibian that is listed as endangered by the Canadian government and is only found on the southernmost inhabited island of Canada, I need to first identify the island and then find the amphibian species that is endangered there. Let's start by identifying the southernmost inhabited island of Canada.\textcolor{brown}{\texttt{</think>}}\\
        \textcolor{teal}{\texttt{<search>}} search query: What is the southernmost inhabited island of Canada?\textcolor{teal}{\texttt{</search>}} \textcolor{blue}{\texttt{<result>}}\\
        \textbf{Extreme points of Canada:} \\
        The southernmost point is Middle Island, in Lake Erie, Ontario (41°41'N 82°40'W); the southernmost water point lies just south of the island, on the Ontario \\
        
        \textbf{Middle Island (Lake Erie):} \\
        Middle Island is a small island, just 18.5 hectares (46 acres) in area. [1] It is the southernmost point of land in Canada, located at 41°41'N, 82°41"W. \\
        
        \textbf{Pelee, Ontario:} \\
        At 42 km2 (16 sq mi), Pelee Island is the largest island in Lake Erie and the southernmost populated point in Canada. An Ontario Historical Plaque was erected\\
        
        \textbf{Ontario:}\\
        Ontario is the southernmost province of Canada. \colorbox{lightgray}{\textit{... Simplified here.}}\\
        
        \textbf{Newfoundland (island):}\\
        Newfoundland is a large island within the Canadian province \colorbox{lightgray}{\textit{... Simplified here.}}\\
        \colorbox{lightgray}{\textit{Similar to R1-Searcher, ReSearch mistakenly identified Middle Island as the southernmost inhabited island}} \\
        \colorbox{lightgray}{\textit{leading to a subsequent incorrect query.}} \\
        \textcolor{blue}{\texttt{</result>}}\textcolor{brown}{\texttt{<think>}} From the information provided, the southernmost inhabited island of Canada is Middle Island in Lake Erie, Ontario. Now, I need to find out which amphibian species is listed as endangered on this island.\textcolor{brown}{\texttt{</think>}}\\
        \textcolor{teal}{\texttt{<search>}} search query: Which amphibian is endangered on Middle Island, Ontario?\textcolor{teal}{\texttt{</search>}} \textcolor{blue}{\texttt{<result>}}\\
        \textbf{List of amphibians of Canada:}\\
        List of amphibians of Canada ; EX - Extinct, EW - Extinct in the Wild ; CR - Critically Endangered, EN - Endangered, VU - Vulnerable ; NT - Near Threatened, LC -\\
        
        \textbf{List of endangered amphibians:}\\
        As of September 2021, the International Union for Conservation of Nature (IUCN) lists 1085 endangered amphibian species.\\

        \textbf{List of Wildlife Species at Risk (Canada):}
        Threatened amphibians · Ambystoma jeffersonianum — Jefferson salamander · Bufo fowleri — Fowler's toad · Desmognathus ochrophaeus — Allegheny Mountain dusky \\

        \colorbox{lightgray}{\textit{... Simplified here. No related information about endangered species on Middle Island.}} \\
        
        \textcolor{blue}{\texttt{</result>}}\textcolor{brown}{\texttt{<think>}} The information provided does not directly mention the amphibian species that is endangered on Middle Island. I need to narrow down the search to amphibians that are specifically listed as endangered in Canada and are found on Middle Island.\textcolor{brown}{\texttt{</think>}}\\
        \textcolor{teal}{\texttt{<search>}} search query: Endangered amphibian on Middle Island, Ontario, Canada\textcolor{teal}{\texttt{</search>}} \textcolor{blue}{\texttt{<result>}}\\
        
        \textbf{List of amphibians of Canada:}\\
        This is a list of amphibians of Canada. Conservation status - IUCN Red List of Threatened Species: EX - Extinct, EW - Extinct in the Wild: CR - Critically\\
        
        \textbf{List of endangered amphibians:}\\
        As of September 2021, the International Union for Conservation of Nature (IUCN) lists 1085 endangered amphibian species.\\
        
        \textbf{Middle Island (Lake Erie):}\\
        Middle Island is a small island, just 18.5 hectares (46 acres) in area. [1] It is the southernmost point of land in Canada, located at 41°41'N, 82°41"W.\\
        
        \colorbox{lightgray}{\textit{... Simplified here. No related information about endangered species on Middle Island.}} \\
        
        \textcolor{blue}{\texttt{</result>}}\textcolor{brown}{\texttt{<think>}} After reviewing the information, I realize that the specific amphibian species that is listed as endangered on Middle Island, Ontario, Canada, is the Jefferson salamander (Ambystoma jeffersonianum). This amphibian is the only one that fits the criteria of being endangered and only found on this specific island.\textcolor{brown}{\texttt{</think>}}\\
        \textcolor{brown}{\texttt{<think>}} Now that I have identified the amphibian species, I can provide the scientific name.\textcolor{brown}{\texttt{</think>}}\\
        <answer> \fbox{Ambystoma jeffersonianum} </answer> \\
        \hline
        \rowcolor{lightgray}\textbf{ReaRAG-9B} \\
        \hline

        \textbf{Thought 1:} So I need to find the scientific name of an endangered amphibian in Canada, and its only population in Canada is on an island that's the southernmost inhabited part of Canada. Okay, first things first, I need to know what the southernmost inhabited island in Canada is. \\
        \textbf{Action 1:} \textcolor{pygreen}{\{\textquotesingle function\textquotesingle : \textquotesingle search\textquotesingle , \textquotesingle parameters\textquotesingle : \{\textquotesingle query\textquotesingle : \textquotesingle What is the southernmost inhabited island in Canada?\textquotesingle \}\}} \\
        \textbf{Observation 1:} The southernmost inhabited island in Canada is Pelee Island, Ontario. \\
        \cline{1-1}
        
        \textbf{Thought 2:} Alright, so the island is Pelee Island in Ontario. Now, I need to find an endangered amphibian that's only found there in Canada. Maybe there are some lists of endangered species in Canada that I can look at. \\
        \textbf{Action 2:} \textcolor{pygreen}{\{\textquotesingle function\textquotesingle : \textquotesingle search\textquotesingle , \textquotesingle parameters\textquotesingle : \{\textquotesingle query\textquotesingle : \textquotesingle List of endangered amphibians in Canada\textquotesingle \}\}} \\
        \textbf{Observation 2:} I am sorry, but I cannot provide a specific list of endangered amphibians in Canada based on the given context. The context provided does not include a list of endangered amphibians in Canada. \\
        \cline{1-1}
        
        \textbf{Thought 3:} Hmm, it seems like I need to find out which amphibians are endangered in Canada and see if any of them are only found on Pelee Island. Maybe I should look for information on endangered species in Ontario, since Pelee Island is in Ontario. \\
        \textbf{Action 3:} \textcolor{pygreen}{\{\textquotesingle function\textquotesingle : \textquotesingle search\textquotesingle , \textquotesingle parameters\textquotesingle : \{\textquotesingle query\textquotesingle : \textquotesingle Endangered amphibians in Ontario\textquotesingle \}\}} \\
        \textbf{Observation 3:} The endangered amphibians in Ontario mentioned in the context are:
1. Jefferson salamander (Ambystoma jeffersonianum) ... 
2. Northern dusky salamander (Desmognathus fuscus) ... \\
\colorbox{lightgray}{\textit{Both species are actually not on Peele Island. ReaRAG later verifies this information.}}\\
        \cline{1-1}

        \textbf{Thought 4:} So, there are a couple of endangered amphibians in Ontario, but I need to find out if any of them are only found on Pelee Island. Maybe I should look into the distribution of these species. \\
        \textbf{Action 4:} \textcolor{pygreen}{\{\textquotesingle function\textquotesingle : \textquotesingle search\textquotesingle , \textquotesingle parameters\textquotesingle : \{\textquotesingle query\textquotesingle : \textquotesingle Distribution of Jefferson salamander in Ontario\textquotesingle \}\}} \\
        \textbf{Observation 4:} The distribution of the Jefferson salamander in Ontario is primarily associated with the Niagara Escarpment and Carolinian forest regions. \\
        \cline{1-1}

        \textbf{Thought 5:} Pelee Island is part of the Lake Erie Islands, which are different from the Niagara Escarpment and Carolinian forest regions. So, perhaps the Jefferson salamander isn't found there. Maybe I should check the distribution of the Northern dusky salamander. \\
        \textbf{Action 5:} \textcolor{pygreen}{\{\textquotesingle function\textquotesingle : \textquotesingle search\textquotesingle , \textquotesingle parameters\textquotesingle : \{\textquotesingle query\textquotesingle : \textquotesingle Distribution of Northern dusky salamander in Ontario\textquotesingle \}\}} \\
        \textbf{Observation 5:} The Northern dusky salamander (Desmognathus fuscus) is restricted to a small area in Ontario. Specifically, it can be found in the Niagara Gorge. \\
        \cline{1-1}

        \textbf{Thought 6:} Again, Niagara Gorge is different from Pelee Island. Maybe there's another endangered amphibian in Ontario that's only found on Pelee Island. Perhaps I need to look for amphibians that are specific to Pelee Island. \\
        \colorbox{lightgray}{\textit{Now, ReaRAG performs a more specific search.}}\\
        \textbf{Action 6:} \textcolor{pygreen}{\{\textquotesingle function\textquotesingle : \textquotesingle search\textquotesingle , \textquotesingle parameters\textquotesingle : \{\textquotesingle query\textquotesingle : \textquotesingle Endangered amphibians specific to Pelee Island, Ontario\textquotesingle \}\}} \\
        \textbf{Observation 6:} Small-mouth salamander. \\
        \cline{1-1}

        \textbf{Thought 7:} So, the small-mouth salamander is found on Pelee Island, but I need to confirm if it's endangered in Canada. \\
        \textbf{Action 7:} \textcolor{pygreen}{\{\textquotesingle function\textquotesingle : \textquotesingle search\textquotesingle , \textquotesingle parameters\textquotesingle : \{\textquotesingle query\textquotesingle : \textquotesingle Is the small-mouth salamander endangered in Canada?\textquotesingle \}\}} \\
        \textbf{Observation 7:} Yes, the small-mouth salamander is listed as endangered in Canada. \\
        \cline{1-1}

        \textbf{Thought 8:} Great, so now I have the scientific name, Ambystoma maculatum, which is the small-mouth salamander. But wait, I need to make sure that it's only found on Pelee Island in Canada. \\
        \textbf{Action 8:} \textcolor{pygreen}{\{\textquotesingle function\textquotesingle : \textquotesingle search\textquotesingle , \textquotesingle parameters\textquotesingle : \{\textquotesingle query\textquotesingle : \textquotesingle Distribution of small-mouth salamander in Canada\textquotesingle \}\}} \\
        \textbf{Observation 8:} The distribution of the small-mouth salamander (Ambystoma texanum) in Canada is restricted solely to Pelee Island. The entire Canadian range of this species is only on Pelee Island. \\
        \cline{1-1}

        \textbf{Thought 9:} Wait a minute, in my earlier steps, I thought it was Ambystoma maculatum, but now the search is showing Ambystoma texanum. There might be some confusion here. \\
        \colorbox{lightgray}{\textit{Despite earlier mistakes, ReaRAG managed to resolve the confusion by taking correct search action}}\\
        \textbf{Action 9:} \textcolor{pygreen}{\{\textquotesingle function\textquotesingle : \textquotesingle search\textquotesingle , \textquotesingle parameters\textquotesingle : \{\textquotesingle query\textquotesingle : \textquotesingle Scientific name of small-mouth salamander\textquotesingle \}\}} \\
        \textbf{Observation 9:} The scientific name of the small-mouth salamander is Ambystoma texanum. \\
        \cline{1-1}
        
        \textbf{Thought 10:} Okay, so it's Ambystoma texanum, not Ambystoma maculatum. I must have misremembered earlier. So, to confirm, Ambystoma texanum is the small-mouth salamander, and it's endangered in Canada, and its only population in Canada is on Pelee Island.\\
        \textbf{Action 10:} \textcolor{pygreen}{\{\textquotesingle function\textquotesingle : \textquotesingle finish\textquotesingle , \textquotesingle parameters\textquotesingle : \{\textquotesingle answer\textquotesingle : \textquotesingle Ambystoma texanum.\textquotesingle \}\}} \\

        \cline{1-1}
        \textbf{Final Answer:} Ambystoma texanum \\
        \hline
        \caption{This case study illustrates the poor generalization of R1-Searcher on FRAMES and FanOutQA. Both R1-Searcher and ReSearch fail to extract the correct information from noisy retrieved documents, mistakenly identifying ``Middle Island'' as Canada's southernmost inhabited area instead of ``Pelee, Ontario'', leading to errors in subsequent steps. In contrast, despite early mistakes, ReaRAG successfully realigns its reasoning path, demonstrating greater robustness.
        \label{tab:poor_generalzation1}}
    \end{longtable}
\end{center}
\newpage

\begin{center}  
    \small
    \begin{longtable}{| p{0.95\linewidth}
                  |}
        \hline
        \rowcolor{lightgray}\textbf{\# Case Study 4: In what year was each member of Blackpink born?}\\
        \hline
        \endfirsthead
        
        \multicolumn{1}{l}{\small \emph{Table~\ref{tab:poor_generalzation1} –
continued from previous page}}\\[0.3em]
        \hline
        \rowcolor{lightgray}
        \multicolumn{1}{| p{0.95\linewidth}
                  |}{\textbf{\# Case Study 4 (continued)}} \\
        \hline
        \endhead
        
        \hline
        \multicolumn{1}{r}{\small \emph{Continued on next page}}\\[-0.6em]
        \endfoot
        
        \endlastfoot
        
        \textbf{Ground truth:} Jisoo - 1995, Jennie - 1996, Rosé - 1997, Lisa - 1997\\
        \hline
        \rowcolor{lightgray}\textbf{R1-Searcher} \\
        \hline
        To solve this question, I need to find out the birthdates of each member of Blackpink. \\

        Step 1: Identify the members of Blackpink.\\
        I'll use a query to search for information about the members of Blackpink:\\
        
        \textcolor{teal}{\texttt{<|begin\_of\_query|>}} Blackpink members\textcolor{teal}{\texttt{<|end\_of\_query|>}}\\
        
        \textcolor{blue}{\texttt{<|begin\_of\_documents|>}}\\
        \textbf{Blackpink:}\\
        The group is composed of four members: Jisoo, Jennie, Rosé, and Lisa. \colorbox{lightgray}{\textit{... Simplified here.}}\\
        \textcolor{blue}{\texttt{<|end\_of\_documents|>}}\\
        
        From the search results, I can see that the members of Blackpink are Jisoo, Jennie, Rosé, and Lisa.\\
        
        Step 2: Find the birthdates of each member.\\
        I'll use a query to search for the birthdates of Jisoo, Jennie, Rosé, and Lisa:\\
        
        \textcolor{teal}{\texttt{<|begin\_of\_query|>}} Jisoo birthdate, Jennie birthdate, Rosé birthdate, Lisa birthdate\textcolor{teal}{\texttt{<|end\_of\_query|>}}\\
        
        \textcolor{blue}{\texttt{<|begin\_of\_documents|>}}\\
        \textbf{Blackpink:}\\
        Blackpink is a South Korean girl group formed by YG Entertainment. \colorbox{lightgray}{\textit{... Simplified here.}} \\
        
        \textbf{Jisoo:}\\
        Jisoo ; Born. Kim Ji-soo. (1995-01-03) January 3, 1995 (age 30). \colorbox{lightgray}{\textit{... Simplified here.}}.\\
        
        \textbf{Rosé (singer):}\\
        Roseanne Park (born 11 February 1997), known mononymously as Rosé, is a New Zealand \colorbox{lightgray}{\textit{... Simplified here.}}\\
        
        \textbf{Jennie (singer)}:\\
        Jennie ; Born. Jennie Kim. (1996-01-16) January 16, 1996 \colorbox{lightgray}{\textit{... Simplified here.}} \\

        \textbf{Blackpink - Simple English Wikipedia, the free encyclopedia:}\\
        The group consists of members Jisoo, Jennie, Rosé, and Lisa. They are the highest-charting female K-pop act on both Billboard Hot 100 and Billboard 200.\\
        
        \textcolor{blue}{\texttt{<|end\_of\_documents|>}}\\
        
        From the search results, I can see the following birthdates:\\
        - Jisoo: January 3, 1995\\
        - Jennie: January 16, 1996\\
        - Rosé: February 11, 1997\\
        - Lisa: Unknown (only age mentioned)\\
        
        Therefore, the year each member of Blackpink was born is:\\
        - Jisoo: 1995\\
        - Jennie: 1996\\
        - Rosé: 1997\\
        - Lisa: Unknown \textcolor{brown}{\texttt{</think>}}\\

        \colorbox{lightgray}{\textit{Although Lisa's birthdate is missing, R1-Searcher proceeds to generate a final answer.}} \\
         <answer> 1995, 1996, 1997, Unknown </answer> \\
        \hline
        \rowcolor{lightgray}\textbf{ReSearch} \\
        \hline
        \textcolor{brown}{\texttt{<think>}} To find the birth years of each member of Blackpink, I will first search for the individual members of the group and then look for their birth years. I will use the Wikipedia search tool to find this information.\textcolor{brown}{\texttt{</think>}}\\
        \textcolor{teal}{\texttt{<search>}} search query: Members of Blackpink\textcolor{teal}{\texttt{</search>}} \textcolor{blue}{\texttt{<result>}}\\
        \textbf{Blackpink:}\\
        The group is composed of four members: Jisoo, Jennie, Rosé, and Lisa. Regarded by various publications as the "biggest girl group in the world", they are\\
        
        \colorbox{lightgray}{\textit{... Simplified here.}}\\
        \textcolor{blue}{\texttt{</result>}}\textcolor{brown}{\texttt{<think>}} Now that I have the names of the members, I will search for their birth years.\textcolor{brown}{\texttt{</think>}}\\
        \textcolor{teal}{\texttt{<search>}} search query: Jisoo birth year\textcolor{teal}{\texttt{</search>}} \textcolor{blue}{\texttt{<result>}}\\
        \textbf{Jisoo:} \\
        Jisoo ; Born. Kim Ji-soo. (1995-01-03) January 3, 1995 (age 30). Gunpo, South Korea ; Education, School of Performing Arts Seoul ; Occupations. Singer; actress.\\
        
        \textbf{Jisoo - Simple English Wikipedia, the free encyclopedia:} \\
        Jisoo ; Born, (1995-01-03) January 3, 1995 (age 30). South Korea ; Occupations. Singer; actress; model ; Origin, Seoul, South Korea ; Genres. K-pop · dance-pop\\
        
        \textbf{Ji Soo:}\\
        Kim Ji-soo (born 30 March 1993), professionally known as Ji Soo, is a South Korean actor currently based in the Philippines.\\
        
        \textbf{Kim Ji-soo (actress, born 1972):}\\
        Kim Ji-soo is a South Korean actress. Kim Ji-soo. Kim Ji-soo at the 5th World Ceramic Biennale press conference in Korea (2009). Born. Yang Sung-yoon.\\
        
        \textbf{Kim Ji-soo (singer, born 1990):}\\
        Kim Ji-soo (singer, born 1990) ; (1990-02-05) February 5, 1990 (age 35) South Korea · Pop; folk · Singer-songwriter; musician.\\
        \textcolor{blue}{\texttt{</result>}}\textcolor{brown}{\texttt{<think>}} I have found the birth year of Jisoo. Now I will search for the birth years of the other members.\textcolor{brown}{\texttt{</think>}}\\
        \textcolor{teal}{\texttt{<search>}} search query: Jennie birth year\textcolor{teal}{\texttt{</search>}} \textcolor{blue}{\texttt{<result>}}\\
        \textbf{Jennie (singer):}\\
        Jennie ; Born. Jennie Kim. (1996-01-16) January 16, 1996 (age 29). Seongnam, Gyeonggi, South Korea ; Other names, Jennie Ruby Jane ; Occupations. Singer; rapper\\
        
        \textbf{Jennie (singer) - Simple English Wikipedia, the free encyclopedia:}\\
        Jennie Kim (born January 16, 1996), known as Jennie  is a South Korean singer and rapper. She is a member of the South Korean girl\\
        
        \textbf{Jennie Baranczyk:}\\
        Jennie Lillis Baranczyk is an American basketball coach who is the current head coach of the University of Oklahoma women's basketball team. Jennie\\
        
        \textbf{Jennie:}\\
        Jennie (film), a 1940 American drama film · Jennie (musical), a 1963 Broadway production · Jennie (novel), a 1994 science fiction thriller by Douglas Preston\\
        
        \textbf{Jennie Garth:}\\
        Jennie Garth ; Garth in 2023 ; Born. Jennifer Eve Garth. (1972-04-03) April 3, 1972 (age 53). Urbana, Illinois, U.S. ; Occupation(s), Actress, fashion designer.\\
        \textcolor{blue}{\texttt{</result>}}\textcolor{brown}{\texttt{<think>}} I have found the birth year of Jennie. Now I will search for the birth years of the other members.\textcolor{brown}{\texttt{</think>}}\\
        \textcolor{teal}{\texttt{<search>}} search query: Rosé birth year\textcolor{teal}{\texttt{</search>}} \textcolor{blue}{\texttt{<result>}}\\
        \textbf{Rosé (singer):}\\
        Rosé. MBE ; Born. Roseanne Park. (1997-02-11) 11 February 1997 (age 28). Auckland, New Zealand ; Other names, Park Chae-young ; Citizenship. South Korea · New\\
        
        \textbf{Rosé (singer) - Simple English Wikipedia, the free encyclopedia:}\\
        Rosé was born in New Zealand on 11 February 1997. At the age of seven, she moved to Australia.  In 2012, she passed a YG Entertainment audition at the age of\\
        
        \textbf{Rose (French singer):}\\
        Keren Meloul (born May 24, 1978), better known as Rose, is a French singer, songwriter, author and composer. Rose. Rose performing in 2008.\\
        
        \textbf{Rosé discography:}\\
        Rosé began her career as a member of girl group Blackpink in August 2016. She released her debut solo single album R through YG Entertainment and Interscope\\
        
        \textbf{Rose Falcon:}\\
        Rose Falcon (born May 2, 1984) is an American singer and songwriter. Falcon's songs have been recorded by Faith Hill, Lady A, Day of Fire, Eric Paslay,\\
        \textcolor{blue}{\texttt{</result>}}\textcolor{brown}{\texttt{<think>}} I have found the birth year of Rosé. Now I will search for the birth year of the last member.\textcolor{brown}{\texttt{</think>}}\\
        \textcolor{teal}{\texttt{<search>}} search query: Lisa birth year\textcolor{teal}{\texttt{</search>}} \textcolor{blue}{\texttt{<result>}}\\
        \textbf{Lisa (rapper):}\\
        Lisa in 2024 ; Born. Pranpriya Manobal. (1997-03-27) March 27, 1997 (age 28). Buriram, Thailand ; Occupations. Rapper; singer; dancer; actress; songwriter.\\
        
        \textbf{Lisa:}\\
        Lisa (Japanese musician, born 1974), stylized "LISA"; Lisa, stagename of Japanese singer Lisa Komine (born 1978); Lisa (South Korean singer) (born 1980); Lisa\\
        
        \textbf{Lisa Ajax:}\\
        Birth name, Lisa Kristina Ajax ; Born, (1998-08-13) 13 August 1998 (age 26) Järfälla, Sweden ; Genres, Pop ; Years active, 2012–present ; Labels, Capitol Records\\
        
        \textbf{Lisa Collier Cool:}\\
        Lisa Collier Cool (born July 6, 1952) is an American journalist and author who writes on health-related topics. Lisa Collier Cool. Born, July 6, 1952.\\
        
        \textbf{Lisa (Japanese musician, born 1987):}\\
        Born. Risa Oribe. (1987-06-24) June 24, 1987 (age 37). Seki, Gifu, Japan ; Occupations. Singer; songwriter ; Years active, 2005–present ; Spouse. Tatsuhisa Suzuki.\\
        \textcolor{blue}{\texttt{</result>}}\textcolor{brown}{\texttt{<think>}} I have found the birth year of Lisa. Now I can compile the birth years of all the members of Blackpink.\textcolor{brown}{\texttt{</think>}}\\
        \textcolor{brown}{\texttt{<think>}} Having found the birth years of each member, I can now provide the answer.\textcolor{brown}{\texttt{</think>}} \\
        \colorbox{lightgray}{\textit{ReSearch fails to reason over previous steps and provide an accurate answer due to noisy information.}} \\
        <answer> \fbox{2018} </answer> \\
        \hline
        \rowcolor{lightgray}\textbf{ReaRAG-9B} \\
        \hline

        \textbf{Thought 1:} I need to find out the birth years of all four members of Blackpink. I should probably look up their names individually to get their birthdates. \\
        \textbf{Action 1:} \textcolor{pygreen}{\{\textquotesingle function\textquotesingle : \textquotesingle search\textquotesingle , \textquotesingle parameters\textquotesingle : \{\textquotesingle query\textquotesingle : \textquotesingle Blackpink members and their birth years\textquotesingle \}\}} \\
        \textbf{Observation 1:} Obs 1: Blackpink members and their birth years are as follows:
        - Jisoo: 1995 \\
        - Jennie: 1996 \\
        - Rosé: 1997 \\
        - Lisa: 1997 \\
        \cline{1-1}
        
        \textbf{Thought 2:} Wait, I should verify this information because sometimes birthdates can be confusing, especially if there are different sources with different dates. \\
        \textbf{Action 2:} \textcolor{pygreen}{\{\textquotesingle function\textquotesingle : \textquotesingle search\textquotesingle , \textquotesingle parameters\textquotesingle : \{\textquotesingle query\textquotesingle : \textquotesingle Confirm birth years of Blackpink members\textquotesingle \}\}} \\
        \textbf{Observation 2:} The birth years of Black Blackpink members are as follows:
        - Jisoo: 1995 \\
        - Jennie: 1996 \\
        - Rosé: The provided in the context \\
        - Lisa: 1997 \\
        \cline{1-1}
        
        \textbf{Thought 3:} Hmm, it seems like there's a gap in the information for Rosé. I need to find her birth year specifically. \\
        \textbf{Action 3:} \textcolor{pygreen}{\{\textquotesingle function\textquotesingle : \textquotesingle search\textquotesingle , \textquotesingle parameters\textquotesingle : \{\textquotesingle query\textquotesingle : \textquotesingle Rosé Blackpink birth year\textquotesingle \}\}} \\
        \textbf{Observation 3:} 1997 \\
        \cline{1-1}

        \textbf{Thought 4:} Okay, so Rosé was born in 1997. Now, I should make sure that all the birth years are correct and consistent across different sources. \\
        \textbf{Action 4:} \textcolor{pygreen}{\{\textquotesingle function\textquotesingle : \textquotesingle search\textquotesingle , \textquotesingle parameters\textquotesingle : \{\textquotesingle query\textquotesingle : \textquotesingle Cross-verify birth years of Blackpink members\textquotesingle \}\}} \\
        \textbf{Observation 4:} The birth years of the Blackpink members are as follows:
        - Jisoo: January 3, 1995 \\
        - Jennie: January 16, 1996 \\
        - Rosé: February 11, 1997 \\
        - Lisa: March 27, 1997 \\
        \cline{1-1}

        \textbf{Thought 5:} Great, now I have the birth years along with the birthdates for each member. This should be sufficient to answer the question.\\
        \textbf{Action 5:} \textcolor{pygreen}{\{\textquotesingle function\textquotesingle : \textquotesingle finish\textquotesingle , \textquotesingle parameters\textquotesingle : \{\textquotesingle answer\textquotesingle : \textquotesingle Jisoo was born in 1995, Jennie in 1996, Rosé in 1997, and Lisa in 1997.\textquotesingle \}\}} \\

        \cline{1-1}
        \textbf{Final Answer:} Jisoo: 1995, Jennie: 1996, Rosé: 1997, Lisa: 1997 \\
        \hline
        \caption{This case study illustrates the poor generalization of R1-Searcher and ReSearch on the FRAMES and FanOutQA benchmarks. Specifically, R1-Searcher fails to resolve missing information about Lisa’s birthdate and does not take additional steps to complete the reasoning process. ReSearch, on the other hand, constructs a noisy reasoning chain, leading to an incorrect answer. In contrast, despite also facing missing information (Rosé’s birthdate), ReaRAG performs additional verification and successfully recovers the correct answer, demonstrating greater robustness.
        \label{tab:poor_generalzation2}}
    \end{longtable}
\end{center}

\begin{table}[H]
    \centering
    \small
    \renewcommand{\arraystretch}{1.3}
    \begin{tabular}{|p{0.95\linewidth}|}
        \hline
        \rowcolor{lightgray}\textbf{\# Case Study 5: In what county is William W. Blair's birthplace located? } \\
        \hline
        \textbf{Ground truth:} Orleans County \\
        \hline
        \rowcolor{lightgray}\textbf{Search-o1} \\
        \hline
        \colorbox{lightgray}{\textit{... Simplified here.}}
        Maybe I can search for "Holley, New York county" to find out which county it's part of.\\\textcolor{teal}{\texttt{<|begin\_search\_query|>}}Holley, New York county\textcolor{teal}{\texttt{<|end\_search\_query|>}}\\\textcolor{blue}{\texttt{<|begin\_search\_result|>}}Holley is located in \textcolor{YellowGreen}{Orleans County, New York.}\textcolor{blue}{\texttt{<|end\_search\_result|>}}\\\colorbox{lightgray}{\textit{... Simplified here.}}
        But wait, let me \textcolor{red}{double-check} to make sure that Holley is indeed in Orleans County. Sometimes place names can be ambiguous, and there might be multiple Holleys in different counties or states.\\Let me search for "Holley, New York county" again to confirm.\\\textcolor{teal}{\texttt{<|begin\_search\_query|>}}Holley, New York county\textcolor{teal}{\texttt{<|end\_search\_query|>}}\\\textcolor{blue}{\texttt{<|begin\_search\_result|>}}You have searched this query. Please refer to previous results.\textcolor{blue}{\texttt{<|end\_search\_result|>}}\\Alright, since the previous search already indicated that Holley is in Orleans County, New York, and no other information contradicts that, I can \textcolor{red}{confidently say that William W. Blair was born in Orleans County.}\\However, to be thorough, perhaps I should \textcolor{red}{verify if there are any other places} named Holley in New York or elsewhere that might be confused with the one where William W. Blair was born.\\Let me search for "Holley, New York" to see if there's only one such place.\\\textcolor{teal}{\texttt{<|begin\_search\_query|>}}Holley, New York\textcolor{teal}{\texttt{<|end\_search\_query|>}}\\\textcolor{blue}{\texttt{<|begin\_search\_result|>}}Holley is a village \colorbox{lightgray}{\textit{... Simplified here.}} \textcolor{blue}{\texttt{<|end\_search\_result|>}}\\
        But just to cover all bases, \textcolor{red}{maybe there's another Holley in New York state} that I should consider. Let me search for "New York state places named Holley" to see if there are multiple locations with that name.\\\textcolor{teal}{\texttt{<|begin\_search\_query|>}}New York state places named Holley\textcolor{teal}{\texttt{<|end\_search\_query|>}}\\\textcolor{blue}{\texttt{<|begin\_search\_result|>}}Holley is a village \colorbox{lightgray}{\textit{... Simplified here.}}\textcolor{blue}{\texttt{<|end\_search\_result|>}} \\
        \colorbox{lightgray}{\textit{Search-o1 continues to execute excessive and unnecessary searches. Only search queries are listed here for simplicity.}} \\
        \textcolor{teal}{\texttt{<|begin\_search\_query|>}}William W. Blair historian\textcolor{teal}{\texttt{<|end\_search\_query|>}}\\
        \textcolor{teal}{\texttt{<|begin\_search\_query|>}}William W. Blair Latter Day Saints\textcolor{teal}{\texttt{<|end\_search\_query|>}}\\
        \textcolor{teal}{\texttt{<|begin\_search\_query|>}}William W. Blair birth certificate Orleans County\textcolor{teal}{\texttt{<|end\_search\_query|>}}\\
        \textcolor{teal}{\texttt{<|begin\_search\_query|>}}William W. Blair biography\textcolor{teal}{\texttt{<|end\_search\_query|>}}
        \\Perfect, so this biography confirms that he was born in Holley, Orleans County, New York. This aligns with all the previous information I've found.\\Therefore, I can confidently say that William W. Blair's birthplace is located in Orleans County.
        \colorbox{lightgray}{\textit{... Simplified here.}}
        \\\textbf{Final Answer}\\
        \fbox{Orleans County}
        
        \\
        \hline
        \rowcolor{lightgray}\textbf{ReaRAG-9B} \\
        \hline

        \textbf{Thought 1:} I need to find out in which county William W. Blair was born. First, I should figure out where he was born, meaning his birthplace city or town. \\
        \textbf{Action 1:} \textcolor{pygreen}{\{\textquotesingle function\textquotesingle : \textquotesingle search\textquotesingle , \textquotesingle parameters\textquotesingle : \{\textquotesingle query\textquotesingle : \textquotesingle Where was William W. Blair born? \textquotesingle \}\}} \\
        \textbf{Observation 1:} William W. Blair was born in Holley, New York. \\
        \cline{1-1}
        
        \textbf{Thought 2:}  Okay, so now I know that his birthplace is Holley, New York. Next, I need to determine which county Holley is located in. \\
        \textbf{Action 2:} \textcolor{pygreen}{\{\textquotesingle function\textquotesingle : \textquotesingle search\textquotesingle , \textquotesingle parameters\textquotesingle : \{\textquotesingle query\textquotesingle : \textquotesingle Which county is Holley, New York in?\textquotesingle \}\}} \\
        \textbf{Observation 2:} Holley, New York is in Orleans County. \\
        \cline{1-1}

        \textbf{Thought 3:}  So, if Holley is in Orleans County, then that should be the answer. But I should double-check if there's any ambiguity or if there are multiple places named Holley in New York. \\
        \textbf{Action 3:} \textcolor{pygreen}{\{\textquotesingle function\textquotesingle : \textquotesingle search\textquotesingle , \textquotesingle parameters\textquotesingle : \{\textquotesingle query\textquotesingle : \textquotesingle Are there multiple Holleys in New York State?\textquotesingle \}\}} \\
        \textbf{Observation 3:} Based on the provided context, there is only one Holley mentioned, which is in Orleans County, New York. The context does not provide information about other Holleys in New York State. Therefore, based on the given context, I cannot confirm whether there are multiple Holleys in New York State. \\
        \cline{1-1}
        
        \textbf{Thought 4:} Given that, and considering the information I have, I think it's safe to conclude that William W. Blair's birthplace, Holley, is in Orleans County, New York.\\
        \textbf{Action 4:} \textcolor{pygreen}{\{\textquotesingle function\textquotesingle : \textquotesingle finish\textquotesingle , \textquotesingle parameters\textquotesingle : \{\textquotesingle answer\textquotesingle : \textquotesingle Orleans County\textquotesingle \}\}} \\

        \cline{1-1}
        \textbf{Final Answer:} Orleans County \\
        \hline
    \end{tabular}
    \caption{We compare the generations of Search-o1 and ReaRAG. Words highlighted in \textcolor{red}{red} indicate instances where Search-o1 overthinks and performs excessive verification. Words highlighted in \textcolor{YellowGreen}{green} show that the correct answer was already found in earlier steps. ReaRAG finds the correct answer in just two searches, whereas Search-o1 requires eleven, generating overly redundant queries that introduce unnecessary computations to multi-hop QA. }
    \label{tab:casestudy_redundant1}
\end{table}

\begin{table}[H]
    \centering
    \small
    \renewcommand{\arraystretch}{1.3}
    \begin{tabular}{|p{0.95\linewidth}|}
        \hline
        \rowcolor{lightgray}\textbf{\# Case Study 6: Where is Ulrich Walter's employer headquartered? } \\
        \hline
        \textbf{Ground truth:} Cologne \\
        \hline
        \rowcolor{lightgray}\textbf{Search-o1} \\
        \hline
        Alright, I have this question: "Where is Ulrich Walter's employer \colorbox{lightgray}{\textit{... Simplified here.}}
        \\\textcolor{teal}{\texttt{<|begin\_search\_query|>}}Who is Ulrich Walter?\textcolor{teal}{\texttt{<|end\_search\_query|>}}\\\textcolor{blue}{\texttt{<|begin\_search\_result|>}}Ulrich Walter is a German astronaut who was selected in 1987 to join the German astronaut team and trained at the German Aerospace Center (DLR). His employer, the German Aerospace Center, is headquartered in Cologne, Germany.\textcolor{blue}{\texttt{<|end\_search\_result|>}}\\Hmm, according to this search result, Ulrich Walter is a German astronaut who was selected in 1987 and trained at the German Aerospace Center (DLR). It also says that his employer, the DLR, is headquartered in Cologne, Germany.\\\colorbox{lightgray}{\textit{... Simplified here.}}
        \textcolor{teal}{\texttt{<|begin\_search\_query|>}}Headquarters of German Aerospace Center\textcolor{teal}{\texttt{<|end\_search\_query|>}}\\\textcolor{blue}{\texttt{<|begin\_search\_result|>}}The German Aerospace Center (DLR) is headquartered in \textcolor{YellowGreen}{Cologne, Germany}.\textcolor{blue}{\texttt{<|end\_search\_result|>}}\\So, according to this, the DLR is headquartered in Cologne, Germany.\\But just to be thorough, \textcolor{red}{let me see if there is any other information} about Ulrich Walter's employer.\\\textcolor{teal}{\texttt{<|begin\_search\_query|>}}Ulrich Walter workplace\textcolor{teal}{\texttt{<|end\_search\_query|>}}\\\textcolor{blue}{\texttt{<|begin\_search\_result|>}}Ulrich Walter's employer is the German Aerospace Center (DLR), headquartered in Cologne, Germany.\textcolor{blue}{\texttt{<|end\_search\_result|>}}\\\colorbox{lightgray}{\textit{... Simplified here.}}
        \\I should probably \textcolor{red}{confirm if Ulrich Walter is still active} or if he has retired. \colorbox{lightgray}{\textit{... Simplified here.}}
        \\\textcolor{teal}{\texttt{<|begin\_search\_query|>}}Ulrich Walter retirement\textcolor{teal}{\texttt{<|end\_search\_query|>}}\\\textcolor{blue}{\texttt{<|begin\_search\_result|>}}No helpful information found.\textcolor{blue}{\texttt{<|end\_search\_result|>}}\colorbox{lightgray}{\textit{... Simplified here.}} \\
        Alternatively, \textcolor{red}{perhaps he has taken up an academic position at a university.}\\\textcolor{teal}{\texttt{<|begin\_search\_query|>}}Ulrich Walter university affiliation\textcolor{teal}{\texttt{<|end\_search\_query|>}}\\\textcolor{blue}{\texttt{<|begin\_search\_result|>}}Ulrich Walter's employer, the German Aerospace Center (DLR), is headquartered in Cologne, Germany.\textcolor{blue}{\texttt{<|end\_search\_result|>}}\\
        \colorbox{lightgray}{\textit{... Simplified here.}}Maybe I should \textcolor{red}{look for his professional profile or LinkedIn page }to see his current employer.\\\textcolor{teal}{\texttt{<|begin\_search\_query|>}}Ulrich Walter LinkedIn\textcolor{teal}{\texttt{<|end\_search\_query|>}}\\
        \colorbox{lightgray}{\textit{... Simplified here.}} the answer to the question "Where is Ulrich Walter's employer headquartered?" is Cologne, Germany. \\\textbf{Final Answer}\\
        \fbox{Cologne, Germany}
        
        \\
        \hline
        \rowcolor{lightgray}\textbf{ReaRAG-9B} \\
        \hline

        \textbf{Thought 1:} I need to find out where Ulrich Walter's employer is headquartered. First, I should figure out who Ulrich Walter is and where he works. \\
        \textbf{Action 1:} \textcolor{pygreen}{\{\textquotesingle function\textquotesingle : \textquotesingle search\textquotesingle , \textquotesingle parameters\textquotesingle : \{\textquotesingle query\textquotesingle : \textquotesingle Who is Ulrich Walter and where does he work?\textquotesingle \}\}} \\
        \textbf{Observation 1:} Ulrich Walter is a German astronaut who worked at the German Aerospace Center. \\
        \cline{1-1}
        
        \textbf{Thought 2:} So, Ulrich Walter works at the German Aerospace Center. Now, I need to find out where this center is headquartered. \\
        \textbf{Action 2:} \textcolor{pygreen}{\{\textquotesingle function\textquotesingle : \textquotesingle search\textquotesingle , \textquotesingle parameters\textquotesingle : \{\textquotesingle query\textquotesingle : \textquotesingle Where is the German Aerospace Center headquartered?\textquotesingle \}\}} \\
        \textbf{Observation 2:} The German Aerospace Center (DLR) is headquartered in Cologne, Germany. \\
        \cline{1-1}
        
        \textbf{Thought 3:} Alright, so if DLR is headquartered in Cologne, Germany, then that should be the answer. But maybe I should double-check if Ulrich Walter works there still or if there's any update. \\
        \textbf{Action 3:} \textcolor{pygreen}{\{\textquotesingle function\textquotesingle : \textquotesingle search\textquotesingle , \textquotesingle parameters\textquotesingle : \{\textquotesingle query\textquotesingle : \textquotesingle Is Ulrich Walter still working at DLR?\textquotesingle \}\}} \\
        \textbf{Observation 3:} I cannot provide real-time information or updates on whether Ulrich Walter is still working at the German Aerospace Center (DLR) since my training data only goes up until 2023. To get the most current information on Ulrich Walter's employment status, you would need to check the latest sources or contact the German Aerospace Center directly. \\
        \cline{1-1}
        
        \textbf{Thought 4:} Given that, I should probably proceed with the information I have, which is that Ulrich Walter works at DLR, which is headquartered in Cologne, Germany.\\
        \textbf{Action 4:} \textcolor{pygreen}{\{\textquotesingle function\textquotesingle : \textquotesingle finish\textquotesingle , \textquotesingle parameters\textquotesingle : \{\textquotesingle answer\textquotesingle : \textquotesingle Cologne, Germany\textquotesingle \}\}} \\

        \cline{1-1}
        \textbf{Final Answer:} Cologne, Germany \\
        \hline
    \end{tabular}
    \caption{We compare the generations of Search-o1 and ReaRAG. Words highlighted in \textcolor{red}{red} indicate instances where Search-o1 overthinks and performs excessive verification. Words highlighted in \textcolor{YellowGreen}{green} show that the correct answer was already found in earlier steps. As shown above, ReaRAG arrives at the correct answer in just three searches, while Search-o1 requires eight, producing excessive queries that are redundant in multi-hop QA. }
    \label{tab:casestudy_redundant2}
\end{table}

\begin{table}[H]
    \centering
    \small
    \renewcommand{\arraystretch}{1.3}
    \begin{tabular}{|p{0.95\linewidth}|}
        \hline
        \rowcolor{lightgray}\textbf{\# Case Study 7: Who is the grandmother of Philippe, Duke of Orléans? } \\
        \hline
        \textbf{Ground truth:} Marie de' Medici \\
        \hline
        \rowcolor{lightgray}\textbf{ReaRAG-9B} \\
        \hline

        \textbf{Thought 1:} So I need to find out who the grandmother of Philippe, Duke of Orléans, is. I don't know much about French royalty, so I'll need to look this up. \\
        \textbf{Action 1:} \textcolor{pygreen}{\{\textquotesingle function\textquotesingle : \textquotesingle search\textquotesingle , \textquotesingle parameters\textquotesingle : \{\textquotesingle query\textquotesingle : \textquotesingle Who is the grandmother of Philippe, Duke of Orléans?\textquotesingle \}\}} \\
        \textbf{Observation 1:} The grandmother of Philippe, Duke of Orléans (Philippe I) is Anne of Austria. \colorbox{lightgray}{\textit{Wrong answer!}}  \\
        \cline{1-1}
        
        \textbf{Thought 2:} Wait, let me make sure about this. I have found that his grandmother is Anne of Austria, but I should verify this information from multiple sources to ensure accuracy. \\
        \textbf{Action 2:} \textcolor{pygreen}{\{\textquotesingle function\textquotesingle : \textquotesingle search\textquotesingle , \textquotesingle parameters\textquotesingle : \{\textquotesingle query\textquotesingle : \textquotesingle Philippe, Duke of Orléans family tree\textquotesingle \}\}} \\
        \textbf{Observation 2:} Philippe, Duke of Orléans, was the younger son of Louis XIII of France and Anne of Austria. His older brother was Louis XIV, famously known as the "Sun King." Philippe became Duke of Orléans upon the death of his uncle Gaston in 1660. In 1661, he also received the dukedoms of Valois and Chartres. \\
        \cline{1-1}
        
        \textbf{Thought 3:} So, according to this, his grandmother is Anne of Austria. But let me see if there's any confusion here. Philippe's parents were Louis XIII and Anne of Austria, \textbf{so she would be his mother, not grandmother}. Wait, that doesn't make sense. Maybe I need to clarify the generational relationships. \colorbox{lightgray}{\textit{ReaRAG notices the mistakes.}}\\
        \textbf{Action 3:} \textcolor{pygreen}{\{\textquotesingle function\textquotesingle : \textquotesingle search\textquotesingle , \textquotesingle parameters\textquotesingle : \{\textquotesingle query\textquotesingle : \textquotesingle Genealogy of the House of Orléans \textquotesingle \}\}} \\
        \textbf{Observation 3:} The House of Orléans is a cadet branch of the House of Bourbon, which is one of the most prominent royal houses in European history. The Orléans branch originated with Philippe I, Duke of Orléans, who was the younger son of Louis XIII of France and Anne of Austria. Philippe I was the brother of Louis XIV, the "Sun King."  The genealogy of the House of Orléans can be traced through various descendants of Philippe I, including:  1. Philippe II, Duke of Orléans, who was the father of Louise Élisabeth d'Orléans mentioned in Document 351. Philippe II was also known as Philippe I's grandson, as he was the son of Philippe I's son, Louis XIV.  2. Louis Philippe, Duke of Orléans, who became King of the French after the July Revolution of 1830. He was the father of François d'Orléans, Prince of Joinville, mentioned in Document 393.  These are just a few examples of the descendants of Philippe I, Duke of Orléans, and the House of Orléans continues to have living members today. \\
        \cline{1-1}

        \textbf{Thought 4:} Hmm, I think I'm getting confused here. Philippe I, Duke of Orléans, was the son of Louis XIII and Anne of Austria. So, Anne of Austria would be his mother, not his grandmother. His grandmother would be the mother of his parents, which would be Louis XIII's mother. Let me check who that was. \colorbox{lightgray}{\textit{ReaRAG realigns its reasoning direction.}}\\
        \textbf{Action 4:} \textcolor{pygreen}{\{\textquotesingle function\textquotesingle : \textquotesingle search\textquotesingle , \textquotesingle parameters\textquotesingle : \{\textquotesingle query\textquotesingle : \textquotesingle Who was the mother of Louis XIII of France? \textquotesingle \}\}} \\
        \textbf{Observation 4:} The mother of Louis XIII of France was Marie de' Medici. \\
        \cline{1-1}

        \textbf{Thought 5:} So, if Marie de' Medici was the mother of Louis XIII, and Anne of Austria was his wife and the mother of Philippe I, then Anne of Austria would be Philippe I's mother, not grandmother. So, his grandmother would be Marie de' Medici. That makes more sense.  But to be thorough, I should confirm the relationships again. Louis XIII married Anne of Austria, and they had children, including Philippe I and Louis XIV. So, Anne of Austria is Philippe I's mother, and Marie de' Medici is his grandmother.  I think that's correct, but I'll double-check with another search. \\
        \textbf{Action 5:} \textcolor{pygreen}{\{\textquotesingle function\textquotesingle : \textquotesingle search\textquotesingle , \textquotesingle parameters\textquotesingle : \{\textquotesingle query\textquotesingle : \textquotesingle Philippe, Duke of Orléans family tree \textquotesingle \}\}} \\
        \textbf{Observation 5:} Philippe, Duke of Orléans (1640-1701) was the younger son of Louis XIII of France and Anne of Austria. His older brother was Louis XIV, the famous "Sun King." Philippe was styled Duke of Anjou from birth and became Duke of Orléans upon the death of his uncle Gaston in 1660. He also received the dukedoms of Valois and Chartres in 1661. Philippe had several children, including Louise Élisabeth d'Orléans, who was one of his seven daughters. Louise Élisabeth was known as Élisabeth and was a "princesse du sang" as a member of the reigning House of Bourbon. \\
        \cline{1-1}
        
        \textbf{Thought 6:} Okay, so this confirms that Anne of Austria was Philippe I's mother. Therefore, his grandmother would be Marie de' Medici, who was Louis XIII's mother. I think that's the correct answer.  But just to be absolutely sure, let's look up the genealogy again. Louis XIII was the son of Henry IV and Marie de' Medici, so Marie de' Medici would be his mother, and Anne of Austria would be his wife and Philippe I's mother.  I think that's consistent across all the information I've found. So, the grandmother of Philippe, Duke of Orléans, is Marie de' Medici.  I'll go ahead and provide that as the answer.\\
        \textbf{Action 6:} \textcolor{pygreen}{\{\textquotesingle function\textquotesingle : \textquotesingle finish\textquotesingle , \textquotesingle parameters\textquotesingle : \{\textquotesingle answer\textquotesingle : \textquotesingle Marie de’ Medici\textquotesingle \}\}} \\

        \cline{1-1}
        \textbf{Final Answer:} Marie de’ Medici \\
        \hline
    \end{tabular}
    \caption{In the first reasoning step, ReaRAG makes an incorrect deduction, mistakenly identifying Anne of Austria as Philippe’s grandmother instead of his mother. However, it quickly recognizes the error, then verifies its previous reasoning steps, and recalibrates its logic. By refining its approach, ReaRAG ultimately arrives at the correct answer, demonstrating its adaptive reasoning ability. }
    \label{tab:casestudy3}
\end{table}

\begin{table}[H]
    \centering
    \small
    \renewcommand{\arraystretch}{1.3}
    \begin{tabular}{|p{0.95\linewidth}|}
        \hline
        \rowcolor{lightgray}\textbf{\# Case Study 8: who is the younger brother of The episode guest stars of The Hard Easy} \\
        \hline
        \textbf{Ground truth:} Bill Murray \\
        \hline
        \rowcolor{lightgray}\textbf{ReaRAG-9B} \\
        \hline

        \textbf{Thought 1:} I need to find out who the younger brother is of the episode guest stars of "The Hard Easy." First, I should figure out what "The Hard Easy" is. It sounds like a TV show or a movie. \\
        \textbf{Action 1:} \textcolor{pygreen}{\{\textquotesingle function\textquotesingle : \textquotesingle search\textquotesingle , \textquotesingle parameters\textquotesingle : \{\textquotesingle query\textquotesingle : \textquotesingle What is The Hard Easy?\textquotesingle \}\}} \\
        \textbf{Observation 1:} The Hard Easy is a 2006 American action crime drama film directed by Ari Ryan and written by Jon Lindstrom and Tom Schanley. It stars Henry Thomas, Vera Farmiga, David Boreanaz, Bruce Dern, Peter Weller, Gary Busey, and Nick Lachey. The film was released straight-to-DVD in the United States on May 22, 2007. \\
        \cline{1-1}
        
        \textbf{Thought 2:} So, "The Hard Easy" is a 2006 film with several guest stars. I need to identify the guest stars and then find out who the younger brother of one of them is. But the question seems a bit unclear. Maybe it's asking for the younger brother of one of the guest stars in a specific episode, but since it's a film, not a TV series, perhaps there are multiple episodes. \\
        \textbf{Action 2:} \textcolor{pygreen}{\{\textquotesingle function\textquotesingle : \textquotesingle search\textquotesingle , \textquotesingle parameters\textquotesingle : \{\textquotesingle query\textquotesingle : \textquotesingle Are there multiple episodes of The Hard Easy?\textquotesingle \}\}} \\
        \textbf{Observation 2:} No, there is only one episode of "The Hard Easy" as part of the "Adventure Time" series. \\
        \cline{1-1}
        
        \textbf{Thought 3:} Wait, now I'm getting confused. Earlier, I thought "The Hard Easy" was a 2006 film, but now it's saying it's an episode of "Adventure Time." Maybe there's some confusion here. I need to clarify what "The Hard Easy" refers to. \\
        \colorbox{lightgray}{\textit{ReaRAG actively performs searches to resolve the confusion.}} \\
        \textbf{Action 3:} \textcolor{pygreen}{\{\textquotesingle function\textquotesingle : \textquotesingle search\textquotesingle , \textquotesingle parameters\textquotesingle : \{\textquotesingle query\textquotesingle : \textquotesingle Is The Hard Easy a film or an episode? \textquotesingle \}\}} \\
        \textbf{Observation 3:} The Hard Easy is both a film and an episode. The film version was released straight-to-DVD in 2007, while the episode is part of the animated television series "Adventure Time" and aired on Cartoon Network in 2012. \\
        \cline{1-1}

        \textbf{Thought 4:} Okay, so there's a 2006 film called "The Hard Easy" and an episode with the same name in the "Adventure Time" series. The question might be referring to the episode in "Adventure Time." I need to find out the guest stars of that episode and then find out who the younger brother of one of them is. \\
        \textbf{Action 4:} \textcolor{pygreen}{\{\textquotesingle function\textquotesingle : \textquotesingle search\textquotesingle , \textquotesingle parameters\textquotesingle : \{\textquotesingle query\textquotesingle : \textquotesingle Who are the guest stars in the Adventure Time episode "The Hard Easy"? \textquotesingle \}\}} \\
        \textbf{Observation 4:} The guest stars in the Adventure Time episode "The Hard Easy" are Brian Doyle-Murray as Prince Huge and Jonathan Katz as the Mudscamp elder. \\
        \cline{1-1}

        \textbf{Thought 5:} So, the guest stars are Brian Doyle-Murray and Jonathan Katz. Now, I need to find out who the younger brother of one of them is. But I don't know their family relationships off the top of my head. I'll need to look that up.\\
        \textbf{Action 5:} \textcolor{pygreen}{\{\textquotesingle function\textquotesingle : \textquotesingle search\textquotesingle , \textquotesingle parameters\textquotesingle : \{\textquotesingle query\textquotesingle : \textquotesingle Who is the younger brother of Brian Doyle-Murray? \textquotesingle \}\}} \\
        \textbf{Observation 5:} The younger brother of Brian Doyle-Murray is Bill Murray. \\
        \cline{1-1}

        \textbf{Thought 6:} Wait, that can't be right. Bill Murray is a famous actor, but I don't think he's the younger brother of Brian Doyle-Murray. Maybe I need to check their relationship again. \\\colorbox{lightgray}{\textit{ReaRAG encounters conflicts with its parametric knowledge.}} \\
        \textbf{Action 6:} \textcolor{pygreen}{\{\textquotesingle function\textquotesingle : \textquotesingle search\textquotesingle , \textquotesingle parameters\textquotesingle : \{\textquotesingle query\textquotesingle : \textquotesingle Are Bill Murray and Brian Doyle-Murray brothers? \textquotesingle \}\}} \\
        \textbf{Observation 6:} Yes, Bill Murray and Brian Doyle-Murray are brothers. \\
        \cline{1-1}
        
        \textbf{Thought 7:} So, if the question is asking for the younger brother of one of the guest stars in the Adventure Time episode "The Hard Easy," and the guest stars are Brian Doyle-Murray and Jonathan Katz, and the younger brother of Brian Doyle-Murray is Bill Murray, then the answer should be Bill Murray.\\
        \textbf{Action 7:} \textcolor{pygreen}{\{\textquotesingle function\textquotesingle : \textquotesingle finish\textquotesingle , \textquotesingle parameters\textquotesingle : \{\textquotesingle answer\textquotesingle : \textquotesingle Bill Murray\textquotesingle \}\}} \\

        \cline{1-1}
        \textbf{Final Answer:} Bill Murray \\
        \hline
    \end{tabular}
    \caption{During the first three iterations, ReaRAG attempts to resolve ambiguity in the original multi-hop question, actively conducting searches to eliminate confusion. Additionally, at the sixth iteration, it encounters knowledge conflicts. Despite these challenges, ReaRAG successfully derives the correct answer, showcasing its robust reasoning abilities. }
    \label{tab:casestudy4}
\end{table}

\begin{table}[H]
    \centering
    \small
    \renewcommand{\arraystretch}{1.3}
    \begin{tabular}{|p{0.95\linewidth}|}
        \hline
        \rowcolor{lightgray}\textbf{\# Case Study 9: This singer of A Rather Blustery Day also voiced what hedgehog?} \\
        \hline
        \textbf{Ground truth:} Sonic \\
        \hline
        
        \rowcolor{lightgray}\textbf{ReaRAG-9B} \\
        \hline

        \textbf{Thought 1:} I need to find out which singer performed "A Rather Blustery Day" and then find out what hedgehog they voiced.\\
        \textbf{Action 1:} \textcolor{pygreen}{\{\textquotesingle function\textquotesingle : \textquotesingle search\textquotesingle , \textquotesingle parameters\textquotesingle : \{\textquotesingle query\textquotesingle : \textquotesingle Who sang "A Rather Blustery Day"?\textquotesingle \}\}} \\
        \textbf{Observation 1:} Jim Cummings sang "A Rather Blustery Day". \\
        \cline{1-1}
        
        \textbf{Thought 2:} Now that I know Jim Cummings sang that song, I need to find out what hedgehog he voiced. \\
        \textbf{Action 2:} \textcolor{pygreen}{\{\textquotesingle function\textquotesingle : \textquotesingle search\textquotesingle , \textquotesingle parameters\textquotesingle : \{\textquotesingle query\textquotesingle : \textquotesingle What hedgehog did Jim Cummings voice?\textquotesingle \}\}} \\
        \textbf{Observation 2:} Jim Cummings voiced Dr. Robotnik from "Sonic the Hedgehog". \\
        \cline{1-1}
        
        \textbf{Thought 3:} Wait, but Dr. Robotnik is a robot, not a hedgehog. Maybe I need to check again.\\
        \textbf{Action 3:} \textcolor{pygreen}{\{\textquotesingle function\textquotesingle : \textquotesingle search\textquotesingle , \textquotesingle parameters\textquotesingle : \{\textquotesingle query\textquotesingle : \textquotesingle What characters did Jim Cummings voice in Sonic the Hedgehog? \textquotesingle \}\}} \\
        \textbf{Observation 3:} Jim Cummings voiced Dr. Robotnik in the "Sonic the Hedgehog" series. \\
        \cline{1-1}

        \textbf{Thought 4:} It seems like there might be confusion here. Dr. Robotnik is a robot, but the question mentions a hedgehog. Maybe the hedgehog is a different character. \\
        \colorbox{lightgray}{\textit{The original question is confusing, but ReaRAG manages to resolve it.}} \\
        \textbf{Action 4:} \textcolor{pygreen}{\{\textquotesingle function\textquotesingle : \textquotesingle search\textquotesingle , \textquotesingle parameters\textquotesingle : \{\textquotesingle query\textquotesingle : \textquotesingle List of characters voiced by Jim Cummings in Sonic the Hedgehog. \textquotesingle \}\}} \\
        \textbf{Observation 4:} Jim Cummings is known for voicing Dr. Robotnik in the "Sonic the Hedgehog" series. \\
        \cline{1-1}

        \textbf{Thought 5:} Maybe I need to look for other hedgehog characters that Jim Cummings has voiced. \\
        \textbf{Action 5:} \textcolor{pygreen}{\{\textquotesingle function\textquotesingle : \textquotesingle search\textquotesingle , \textquotesingle parameters\textquotesingle : \{\textquotesingle query\textquotesingle : \textquotesingle Hedgehog characters voiced by Jim Cummings. \textquotesingle \}\}} \\
        \textbf{Observation 5:} Jim Cummings is known for voicing Dr. Robotnik from "Sonic the Hedgehog". \\
        \cline{1-1}

        \textbf{Thought 6:} Perhaps the hedgehog in question is Sonic himself, who is a hedgehog, but I already know that Sonic is voiced by Roger Craig Smith, not Jim Cummings. \\
        \colorbox{lightgray}{\textit{ReaRAG encounters knowledge conflicts but mitigates them through verification. }} \\
        \textbf{Action 6:} \textcolor{pygreen}{\{\textquotesingle function\textquotesingle : \textquotesingle search\textquotesingle , \textquotesingle parameters\textquotesingle : \{\textquotesingle query\textquotesingle : \textquotesingle Who voices Sonic in the Sonic the Hedgehog series?\textquotesingle \}\}} \\
        \textbf{Observation 6:} Jim Cummings voices Sonic the Hedgehog in the Sonic the Hedgehog series. \\
        \cline{1-1}

        \textbf{Thought 7:} Wait, that contradicts what I thought earlier. I might be confusing the voice actors. \\
        \textbf{Action 7:} \textcolor{pygreen}{\{\textquotesingle function\textquotesingle : \textquotesingle search\textquotesingle , \textquotesingle parameters\textquotesingle : \{\textquotesingle query\textquotesingle : \textquotesingle Voice actor for Sonic the Hedgehog. \textquotesingle \}\}} \\
        \textbf{Observation 7:} Jim Cummings \\
        \cline{1-1}

        \textbf{Thought 8:} This is getting confusing. I need to verify the voice actor for Sonic the Hedgehog. \\
        \textbf{Action 8:} \textcolor{pygreen}{\{\textquotesingle function\textquotesingle : \textquotesingle search\textquotesingle , \textquotesingle parameters\textquotesingle : \{\textquotesingle query\textquotesingle : \textquotesingle Who voices Sonic in the Sonic the Hedgehog games and cartoons? \textquotesingle \}\}} \\
        \textbf{Observation 8:} Jim Cummings voices Sonic the Hedgehog in the Sonic the Hedgehog games and cartoons. \\
        \cline{1-1}
        
        \textbf{Thought 9:} So, Jim Cummings voices Sonic the Hedgehog, who is a hedgehog. Therefore, the answer should be Sonic the Hedgehog. \\
        \textbf{Action 9:} \textcolor{pygreen}{\{\textquotesingle function\textquotesingle : \textquotesingle finish\textquotesingle , \textquotesingle parameters\textquotesingle : \{\textquotesingle answer\textquotesingle : \textquotesingle Sonic the Hedgehog\textquotesingle \}\}} \\

        \cline{1-1}
        \textbf{Final Answer:} Sonic the Hedgehog \\
        \hline
    \end{tabular}
    \caption{This example illustrates how ReaRAG successfully resolves ambiguity in the original question and addresses conflicts between retrieved knowledge and its parametric knowledge. Despite these challenges, ReaRAG effectively mitigates them and remains robust throughout the reasoning process.}
    \label{tab:casestudy5}
\end{table}

\end{document}